\documentclass[10pt,twocolumn,letterpaper]{article}

\usepackage[pagenumbers]{cvpr} 

\definecolor{cvprblue}{rgb}{0.21,0.49,0.74}

\definecolor{best}{HTML}{8FDC97}
\definecolor{second}{HTML}{e3f7d3}
\usepackage[breaklinks,colorlinks,allcolors=cvprblue]{hyperref}

\usepackage{url}
\usepackage{booktabs}       
\usepackage{amsfonts}       
\usepackage{nicefrac}       
\usepackage{microtype}      
\usepackage{xcolor}         
\usepackage{xpatch}
\usepackage{amsmath} 
\usepackage{hyphenat} 
\usepackage{multirow}
\usepackage{makecell}
\usepackage{siunitx}
\usepackage{graphicx}
\usepackage{subcaption} 
\usepackage{caption}
\usepackage{amssymb} 
\usepackage{float}
\usepackage{enumitem}
\usepackage{algorithm}        
\usepackage{algpseudocode}    
\usepackage{amsthm}
\usepackage{tikz}
\usepackage{float}
\usepackage{placeins}   
\usepackage{ragged2e} 
\usepackage{tabularx} 
\usepackage{selinput}
\usepackage[table]{xcolor}
\usepackage{xspace}
\usepackage{wrapfig}

\def\paperID{} 
\def\confName{CVPR}
\def\confYear{2026}

\title{Test-time Sparsity for Extreme Fast Action Diffusion}

\author{
    Kangye Ji$^{1}$\thanks{Equal contribution. Work done when Jianbo Zhou was an intern at mmlab@SIGS, Tsinghua University.},\ 
    Yuan Meng$^{1}$\footnotemark[1]\hspace{0.35em}\thanks{Corresponding authors.},\ 
    Jianbo Zhou$^{1}$\footnotemark[1],\ 
    Ye Li$^{1}$,\ 
    Chen Tang$^{2}$, \
    Zhi Wang$^{1}$\footnotemark[2] \\
    $^{1}$Tsinghua University \quad
    $^{2}$The Chinese University of Hong Kong \\
}

\begin{document}
\maketitle

\begin{abstract}
Action diffusion excels at high-fidelity action generation but incurs heavy computational costs owing to its iterative denoising nature. Despite current technologies showing promise in accelerating diffusion transformers by reusing the cached features, they struggle to adapt to policy dynamics arising from diverse perceptions and multi-round rollout iterations in open environments. 
We propose test-time sparsity to tackle this challenge, which aims to accelerate action diffusion by dynamically predicting prunable residual computations for each model forward at test time.
However, two bottlenecks remain in this paradigm: 1) repetitive conditional encoding and pruning offset most potential speed gains, and 2) the features cached from previous denoising timesteps cannot constrain large pruning errors under aggressive sparsity. To address the first bottleneck, we design a highly parallelized inference pipeline that minimizes the non-decoder delay to milliseconds. Specifically, we first design a lightweight pruner that shares the encoder with the diffusion transformer. Then, we decouple the encoding and pruning from the autoregressive denoising loop by processing all denoising timesteps in parallel, and overlap the pruner with the decoder forward inference through asynchronism. 
To overcome the second bottleneck, we introduce an omnidirectional reusing strategy, which achieves 95\% sparsity by selectively reusing the features cached from the current forward, previous denoising timesteps, and earlier rollout iterations. To learn the rollout-level reusing strategies, we sample a few action trajectories to supervise the sparsified diffusion step by step.
Extensive experiments demonstrate that our method reduces FLOPs by 92\% and accelerates action generation by 5$\times$, achieving lossless performance with an inference frequency of 47.5 Hz. Our code is available at \url{https://github.com/ky-ji/Test-time-Sparsity}.


\end{abstract}

\vspace{-1 em}
    
\section{Introduction}
\label{sec:intro}





\begin{figure}[htbp]
    \centering
    \includegraphics[width=0.5\textwidth]{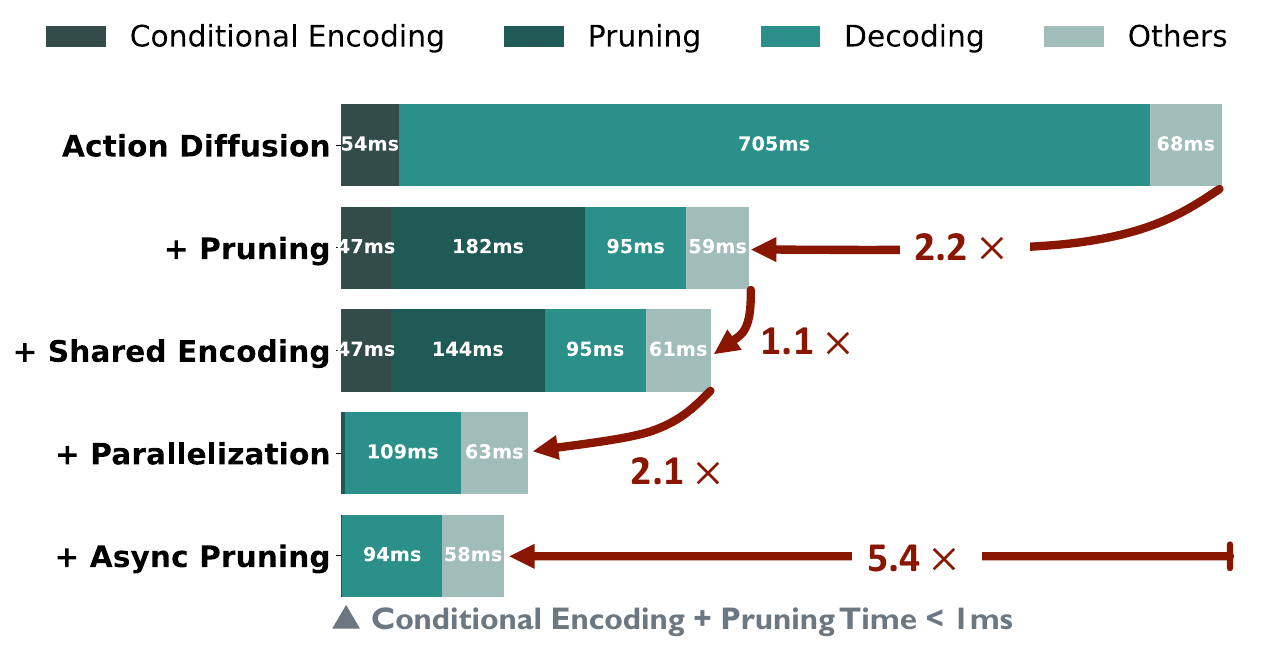}
    \caption{Inference Latency Breakdown. Test-time sparsity removes 5\% of the decoder computations by omnidirectionally reusing historical features (Section \ref{sec: omnidirectional reusing}), and reduces the delay of conditional encoding and pruning to within one millisecond through a highly parallelized inference pipeline (Section~\ref{sec:parallelized pipeline}).}
    \label{fig:inference}
\end{figure}





Action diffusion~\cite{pearce2023imitating,chi2023diffusion,team2024octo} has garnered substantial attention in robotic control due to its remarkable ability to model multi-modal action distributions~\citep{chi2023diffusion, ze20243dp, liu2024rdt,liu2025spatialpolicy, li2025ts}. This capability has made it essential for action generation modules in recent visuomotor policies~\cite {chi2023diffusion,ze20243dp} and advanced Vision-Language-Action (VLA) models~\citep{black2024pi_0,shukor2025smolvla,wen2025dexvla,liu2025hybridvla,hou2025dita,li2025spvla}, particularly for complex and highly dexterous manipulation tasks. However, the adoption of action diffusion is fundamentally constrained by its low achievable frequency inherent in the iterative denoising process, e.g., 6Hz for Diffusion Policy~\cite{chi2023diffusion} and 5Hz for 3D Diffusion Policy~\cite {ze20243dp} on a consumer-grade GPU, far below the 30 Hz required by many real-world tasks~\citep{ma2025runningvlasrealtimespeed}. 



Despite current technologies showing promise in accelerating diffusion transformers by reusing the cached features, they struggle to adapt to the evolving dynamics of visuomotor polices in open environments.
A line of works \cite{hoeg2024streaming, chen2025falcon} accelerate the diffusion by reusing the previous partially denoised actions from earlier rollout iterations, while another line of studies ~\cite{ji2025block,yang2025efficientvla} reuse the intermediate features from the previous denoising steps. 
However, these approaches rely on static schedules, which are inherently mismatched with the dynamic nature of the environment. In other words, they fail to account for the evolving sparsity patterns arising from the multi-round interactions and diverse perceptions that emerge in open, ever-changing environments.




To address this gap, we propose test-time sparsity for extreme fast action diffusion, aiming to predict prunable residual computations before each model forward. The paradigm follows a prune-then-reuse pipeline \cite{ji2026sparse}, where a parameterized pruner dynamically predicts the skippable residual computations of the forward, and then the action diffusion transformer skips the sparse residuals, reducing the pruning error by reusing the features cached from previous timesteps.
However, such a paradigm raises two primary inference bottlenecks that hinder its practical efficiency: 1) repetitive conditional encoding and pruning offset most speed gains (as shown in Figure~\ref{fig:inference}), 2) the features cached from timesteps are insufficient to constrain large pruning errors under aggressive sparsity (refer to Table~\ref{tab: ablation}). 


To address the first bottleneck, we design a highly parallelized inference pipeline that minimizes the non-decoder delay to milliseconds. Specifically, we first design a lightweight parameterized pruner that shares the encoder with the action diffusion transformer, which reduces a 40ms delay, as shown in Figure~\ref{fig:inference}. Then, we decouple the encoding and pruning from the autoregressive denoising loop by processing all denoising timesteps in parallel at test time. Finally, the pruner is overlapped with the decoder forward inference through asynchronism, which reduces the non-decoder delay to milliseconds.


To overcome the second bottleneck, we introduce omnidirectional reusing, which achieves 95\% sparsity by efficiently reusing historical features cached from multiple directions. The insight stems from the observation that features cached from different directions are not only highly aligned with the anchored ones but also offer complementary advantages such as different angles and shorter distances, as illustrated in Figure~\ref{fig:PCA}.
To operationalize this insight, we first model the historical feature space as a 3D lattice, which efficiently organizes the cached features and provides three distinct candidates for each anchored feature. On top of this structure, we design the pruner output as a unified embedding that jointly decides whether to perform computation and which cached feature to reuse.
To overcome the limitations of per-forward supervision in learning rollout-level reusing strategies, we sample the action trajectories to supervise the action generated by the sparsified diffusion step by step. 

To assess the effectiveness of the proposed method, we conduct experiments on extensive benchmarks. Our approach prunes up to 95\% of computations during action diffusion across the entire rollout, yielding 5$\times$ wall-clock speedup and achieving a 47.5 Hz inference frequency on an NVIDIA 4090 GPU without any performance degradation. In summary, our main contributions are threefold:

\begin{figure*}[htbp]
    \centering
    \includegraphics[width=\textwidth]{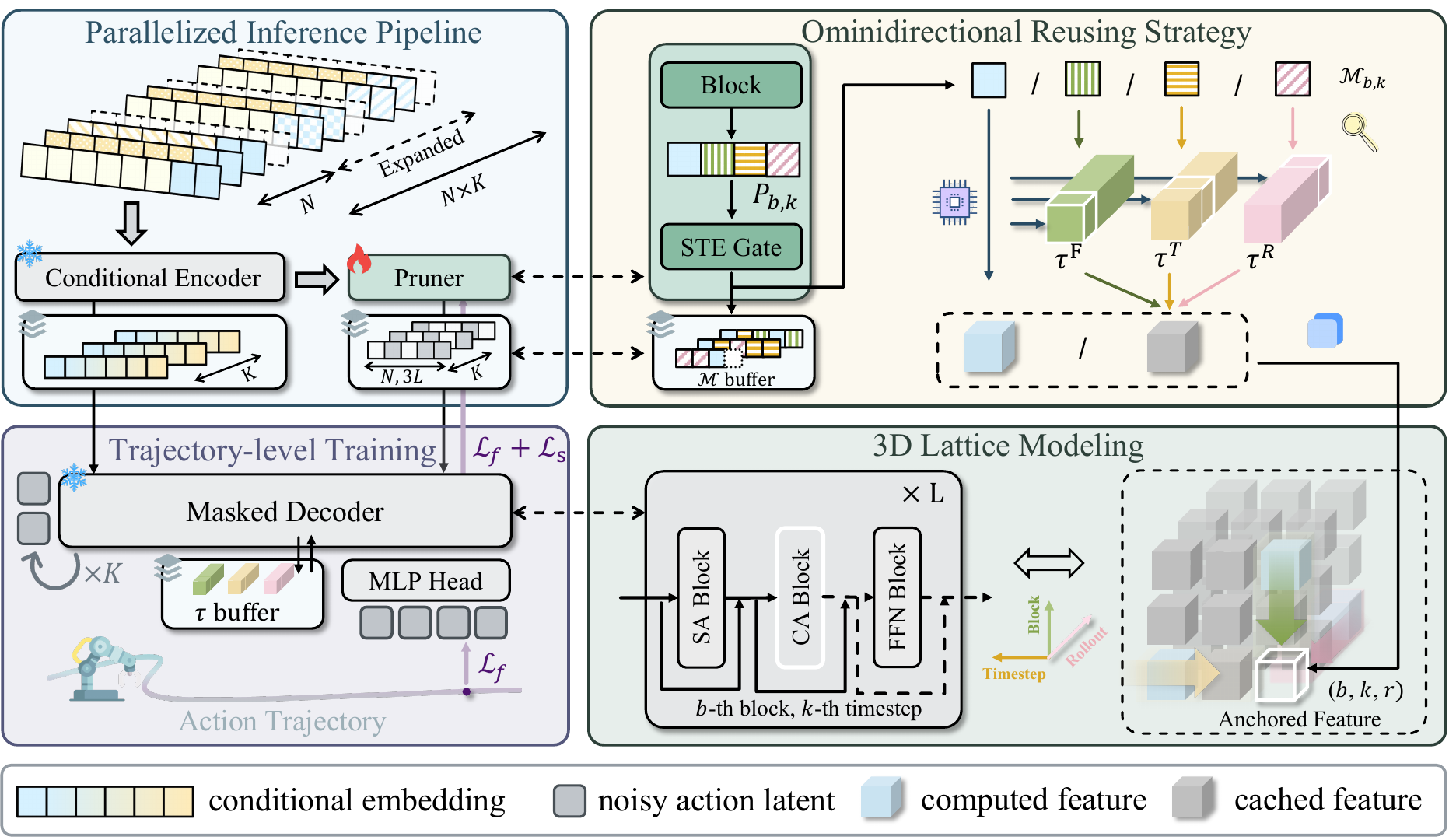}
    \caption{
    The overview of Test-time Sparsity. The paradigm comprises four key components: 1) Parallelized Inference Pipeline that reduces non-decoder latency to milliseconds; 2) Omnidirectional Reusing which achieves extreme sparsity by adaptively reusing features cached from the current forward, previous denoising timesteps, and earlier rollout iterations; 3) 3D Lattice Modeling for efficiently organizing large-scale historical features; and 4) Trajectory-level Training that supervises the learning of Omnidirectional Reusing strategy.
    }
    \label{fig:overview}
\end{figure*}

\begin{itemize}
\item [1.] We propose test-time sparsity for accelerating action diffusion, which dynamically prunes the diffusion transformer to accommodate evolving policy dynamics.

\item [2.] We unleash the instantiation potential of test-sparsity by two key designs: a highly parallelized inference pipeline that reduces the non-decoder overhead to milliseconds, and an omnidirectional reusing strategy that achieves 95\% sparsity by selectively reusing the features cached from the current forward, previous denosing timesteps, and earlier rollout iterations.

\item [3.] We conduct comprehensive experiments on diverse manipulation benchmarks, covering both visuomotor polices and VLA models. The results show that our method achieves 5$\times$ speedup while maintaining high generation fidelity.
\end{itemize}

\section{Related Work}


\paragraph{Action Diffusion Acceleration.}
Current acceleration techniques for action diffusion broadly fall into (i) distillation-based model compression and (ii) reuse-based computation sparsity. 
Distillation-based methods learn fast surrogates that approximate the original multi-step denoising trajectory. One-Step Diffusion Policy~\citep{wang2024onedp} converts a pretrained policy into a single-step action generator. Consistency Policy~\citep{prasad2024consistencydp} and VDD~\citep{zhou2024vdd} distill a Diffusion Policy into a faster student policy by enforcing self-consistency along diffusion trajectories. 
In terms of reuse-based methods, both Falcon~\citep{chen2025falcon} and Streaming Policy~\cite{hoeg2024streaming} reuse partially denoised trajectories in the previous iterations. Other methods reuse the intermediate features. EfficientVLA~\citep{yang2025efficientvla} adopts a uniform reusing schedule, i.e., updating the cached features at fixed intervals. Block-wise Adaptive Caching~\citep{ji2025block} accelerates policy inference by reusing intermediate action features at the block level. However, these methods can not adjust the schedules dynamically according to policy dynamics. 



\paragraph{Reuse-based Image Diffusion Acceleration.}
Previous works explored the sparsity of diffusion models well by skipping redundant activations with cached features. Early methods~\citep{ma2024deepcache,liu2025teacache} focus on U-Net backbones, while more recent approaches~\citep{selvaraju2024fora,ma2024learningtocache,chen2024delta,zou2025accelerating,liu2025fastcache} focus on transformer architectures. 
Due to the diffusion models typically undergoing a one-shot generation process, these methods primarily adopt predefined strategies to reuse historical features from previous timesteps. Most of them identify and prune the most relevant features based on heuristics, such as temporal proximity~\citep{liu2025teacache} or feature similarity~\citep{selvaraju2024fora}. Other works design caching strategies~\citep{chen2024delta} tailored to the specific patterns of the image generation process. Notably, despite some works~\citep{ma2024learningtocache,zhu2024dipgo} learn the sparsity pattern by aligning the action before and after sparsification, they still generate the learned schedules offline to avoid the additional overhead.

\section{Methodology}
\label{sec: method}

In this section, we first outline preliminaries in action diffusion in Section~\ref{subsection: Preliminaries}. Next, we introduce the parallelized inference pipeline in Section~\ref{sec:parallelized pipeline} and describe the omnidirectional reusing strategy in Section~\ref{sec: omnidirectional reusing}.

\subsection{Preliminary}
\label{subsection: Preliminaries}

\paragraph{Action Diffusion Transformer.}

The action diffusion transformer consists of a conditional encoder and a transformer decoder. At each denoising timestep $k$, the conditional encoder maps the diffusion timestep into a sinusoidal embedding and transforms the current observation $\mathbf{o}_r$ into an embedding sequence via an MLP. These embeddings are then concatenated with learnable positional encodings to form the conditional representation. The transformer decoder takes the noisy action $\mathbf{a}_r^k$ as input tokens and refines it through $L$ stacked layers. Each layer $l$ comprises a self-attention (SA) block, a cross-attention (CA) block, and a feed-forward network (FFN). The conditional embeddings are injected into the decoder via the cross-attention blocks, enabling effective conditioning across timesteps.
Given the hidden state $\mathbf{h}_k^{l-1}$ at denoising step $k$, the update is a residual summation:
\begin{equation}
\mathbf{h}_k^{l}
= \mathbf{h}_k^{l-1}
+ \text{SA}_k^{l}
+ \text{CA}_k^{l}
+ \text{FFN}_k^{l}.
\label{eq:dit}
\end{equation}
The output of layer $l$ is computed by summing the residual outputs of these blocks. As the block is the minimal residual computation unit, it serves as the pruning target in this work.

\paragraph{Action Diffusion.}
The action diffusion describes the visuomotor generation process in a diffusion manner during the policy-environment interaction. At each iteration $r$, the policy $\pi$ receives the current observation $\mathbf{o}_r$ and produces an action chunk $\mathbf{a}_r$, which is executed by the robot. The environment then evolves under its dynamics and yields the next observation $\mathbf{o}_{r+1}$:
\begin{equation}
\mathbf{o}_{r+1} \sim P\!\left(\mathbf{o}_{r+1} \mid \mathbf{o}_r, \mathbf{a}_r \right),
\label{eq:env}
\end{equation}
where $P(\cdot \mid \mathbf{o}_r, \mathbf{a}_r)$ denotes the environment’s state transition probability. This process repeats for multiple iterations.

The action $\mathbf{a}_r$ used in each iteration is obtained via conditional denoising diffusion~\citep{chi2023diffusion}. Starting from Gaussian noise $\mathbf{a}_r^{K} \sim \mathcal{N}(0,I)$, the policy performs $K$ reverse denoising steps using the denoiser $f_\theta$:
\begin{equation}
\mathbf{a}_r^{k-1} = f_\theta\!\left(\mathbf{a}_r^{k}, \, \mathbf{o}_{r}, \, k \right),
\quad k = K, \dots, 1.
\label{eq:diffusion}
\end{equation}
We can therefore formulate the entire multi-step denoising process, which maps initial noise to a final action:
\begin{equation}
\mathbf{a}_r = \pi(\mathbf{o}_r) = \mathbf{a}_r^{0} =f_\theta \circ f_\theta \circ \dots \circ f_\theta (\mathbf{a}_r^{K}, \mathbf{o}_r).
\label{eq:full_policy}
\end{equation}
The final $\mathbf{a}_r $ serves as the control action in Eq.~\ref{eq:env}.

\subsection{Parallelized Inference Pipeline}
\label{sec:parallelized pipeline}

\begin{figure}[t]
    \centering
    \includegraphics[width=0.5\textwidth]{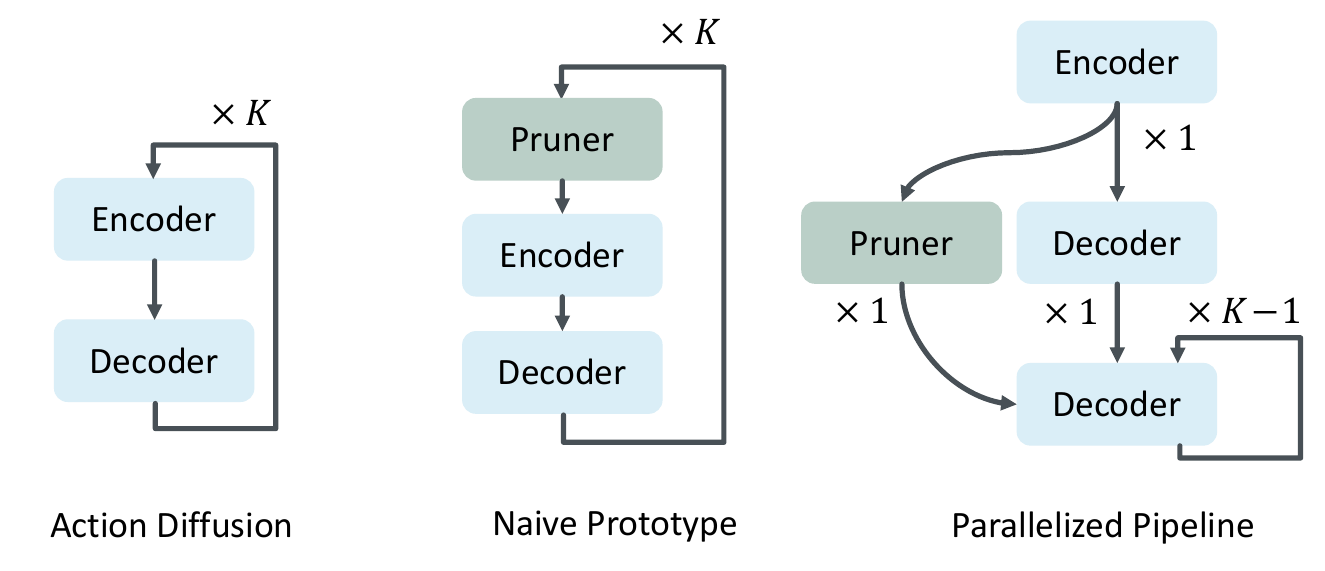}
    \caption{Comparison of different inference pipelines. The parallelized pipeline runs the pruner and encoder only once, and overlaps the runtime of the pruner with the initial decoder step. }
    \label{fig:pipeline}
\end{figure}

\paragraph{Prototype of Test-time Sparsity.}

To adapt to changing dynamics of visuomotor polices in an open environment, we necessitate a pruner that identifies the skippable residual blocks for each forward inference adaptively. We condition the pruner with environment perceptions $\mathbf{o}_r$ and the current denoising timestep $k$, which causally govern the sparsity pattern of each forward process. The pruner is further expected to output a binary pruning mask $\mathcal{M} \in \{0, 1\}^{3L}$ for all $3L$ DiT blocks. The prototype is illustrated in Figure~\ref{fig:pipeline}.

\paragraph{Decoupled Pruner by Parallelization.}

The primary inference bottleneck in the prototype arises from repetitive operations within the autoregressive denoising loop, including decoding, conditional encoding, and pruning. As shown in Figure~\ref{fig:inference}, pruning at test-time has addressed the repetitive decoding well by skipping and reusing previously cached features, which reduces the corresponding delay from 705ms to 95ms in each diffusion. However, this solution introduces an additional significant overhead: the pruner itself requires 182ms for its own repetitive inference, a cost that even surpasses the 95ms decoding time achieved at 95\% sparsity.

\begin{figure}[t]
    \centering 

    \begin{subfigure}[b]{0.23\textwidth}
        \centering
        \includegraphics[width=\linewidth]{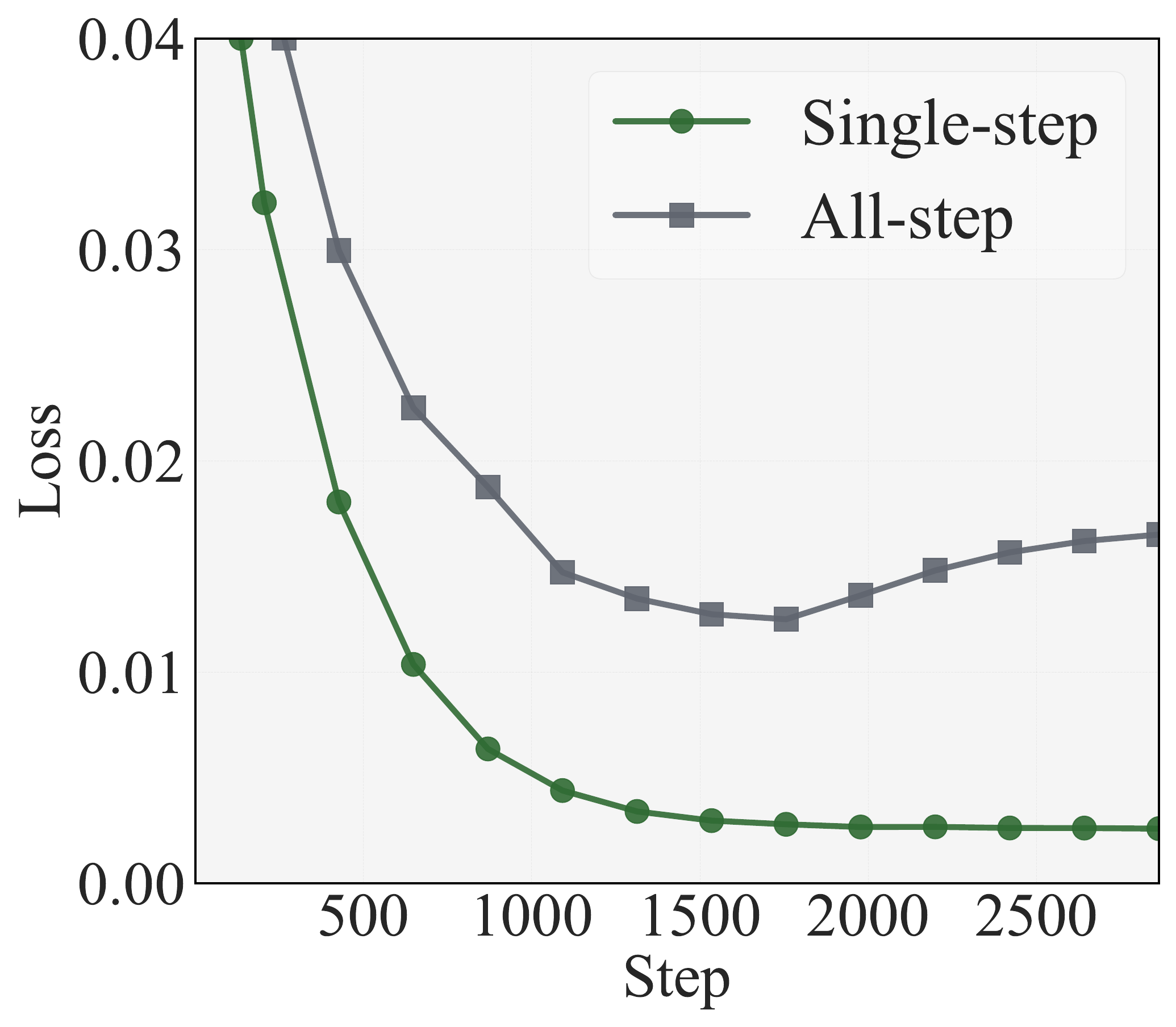}
        \caption{Can task.}
    \end{subfigure}
    \hfill 
    \begin{subfigure}[b]{0.23\textwidth}
        \centering
        \includegraphics[width=\linewidth]{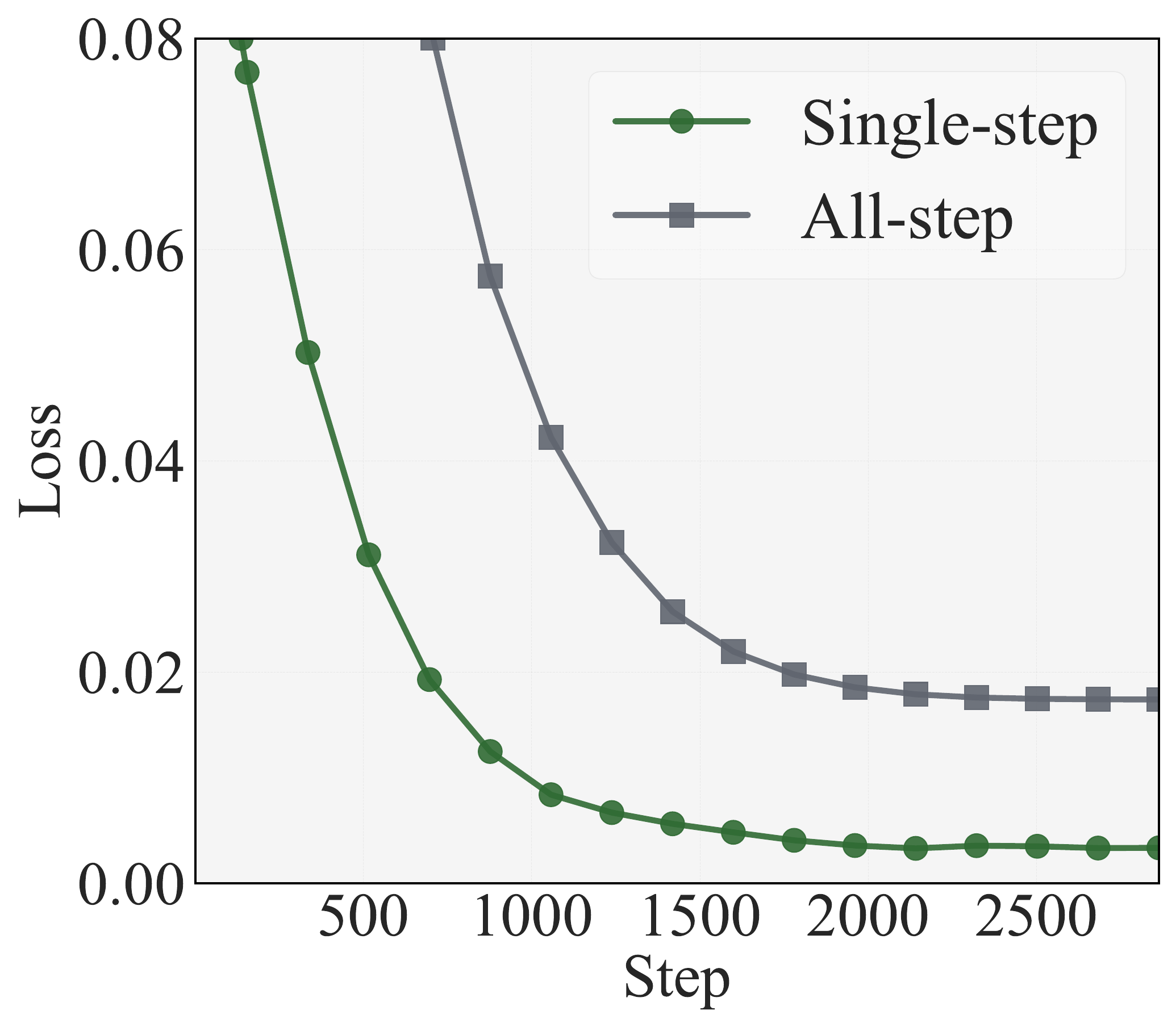}
        \caption{Square task.}
    \end{subfigure}

    \caption{Loss convergence with different timestep encodings.}
    \label{fig:loss convergence}
\end{figure}

To reduce the significant inference overhead of the pruner, a naive idea is to decouple the pruner from the iterative loop by encoding all the timesteps in one go. 
However, as evidenced by the higher curve in Figure~\ref{fig:loss convergence}, the pruner performs substantially worse under all-step encoding, revealing a fundamental trade-off between the efficiency of all-step encoding and the accuracy attainable with single-step encoding. 


To address this dilemma, we parallelize the prediction of the pruner by fully exploiting the GPU at inference time. Although the pruner model is designed to process a single timestep $k$, we reformulate this as a large-scale batched operation at inference. Specifically, we first compute the sinusoidal positional embeddings for all $K$ denoising timesteps simultaneously. We then collapse the temporal dimension $K$ into the data batch dimension $N$, creating a single, large batch with an effective size of $N \times K$, as illustrated in Figure~\ref{fig:overview}. This unified tensor is fed into the pruner in a single forward pass, generating pruning masks for all timesteps concurrently. This parallel formulation transforms the $K$-step sequential loop into a fully-batched operation, thereby eliminating most of the pruner-induced delay.




\paragraph{Lightweight Pruner Design.}
 We instantiate the pruner with a lightweight transformer decoder block. To further reduce the latency of conditional encoding, we share the condition encoder between the pruner and the action diffusion transformer, enabling the pruner to reuse high-level condition embeddings without incurring additional cost. Moreover, by encoding all timesteps in parallel, the conditional encoder not only significantly reduces inference delay but also seamlessly supports the parallelized pruner.

Concretely, the pruner first encodes all the block indices using sinusoidal positional embeddings to obtain block-specific query embeddings. The pruner then processes these block embeddings as the target sequence within its transformer decoder block, which is conditioned on the output of the shared condition encoder from the action diffusion transformer. The decoder outputs are subsequently passed through an MLP head to produce the pruning mask $\mathcal{M}$. 


\paragraph{Asynchronous Pipeline.} 

\begin{figure}[t]
    \centering
    \includegraphics[width=0.48\textwidth]{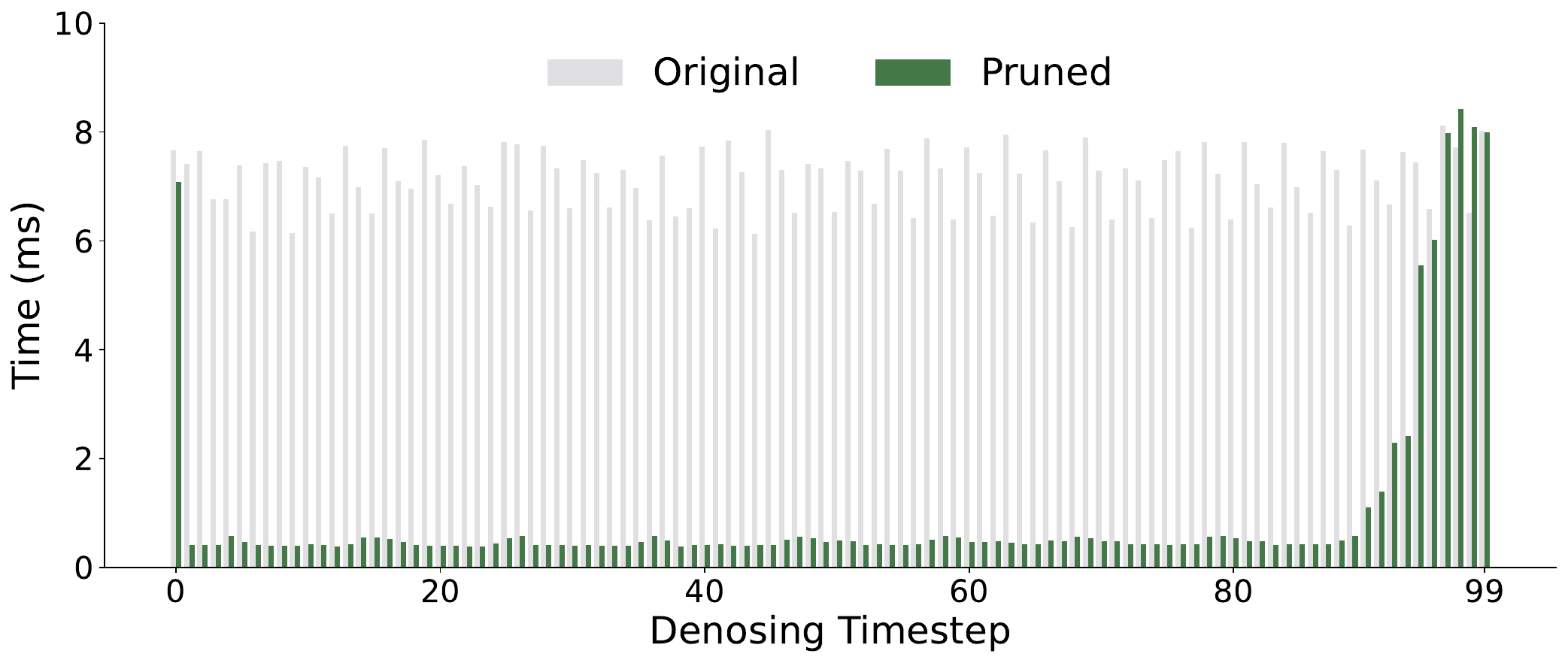}
    \caption{Inference time breakdown during denoising. The inference time after pruning is mainly concentrated in the later steps.}
    \label{fig:diffusion time}
\end{figure}

To facilitate efficient coordination within the decoupled paradigm, we design an asynchronous pipeline. Specifically, two buffers are introduced for the conditional encoder and the pruner to store the condition embeddings and pruning masks generated at once. Afterward, the pipeline transitions into a decoder-only loop, where the decoder directly retrieves these cached features for conditioning and masking, thereby eliminating sequential dependencies and waiting time.

We further analyze the inference time during a single diffusion process and observe in Figure~\ref{fig:diffusion time} that the pruner skips nearly all redundant computations in the early steps, except for the first step, due to the necessary full forward for cache initialization. To hide the runtime of the pruner, we launch two parallel threads for the pruner and decoder after conditional encoding, as illustrated in Figure~\ref{fig:pipeline}, enabling only 0.45ms overhead for the pruning at test time.

\subsection{Ominidirectional Reusing Strategy}
\label{sec: omnidirectional reusing}
\paragraph{Motivation.}

\begin{figure}[htbp]
    \centering 

    \begin{subfigure}[b]{0.23\textwidth}
        \centering
        \includegraphics[width=\linewidth]{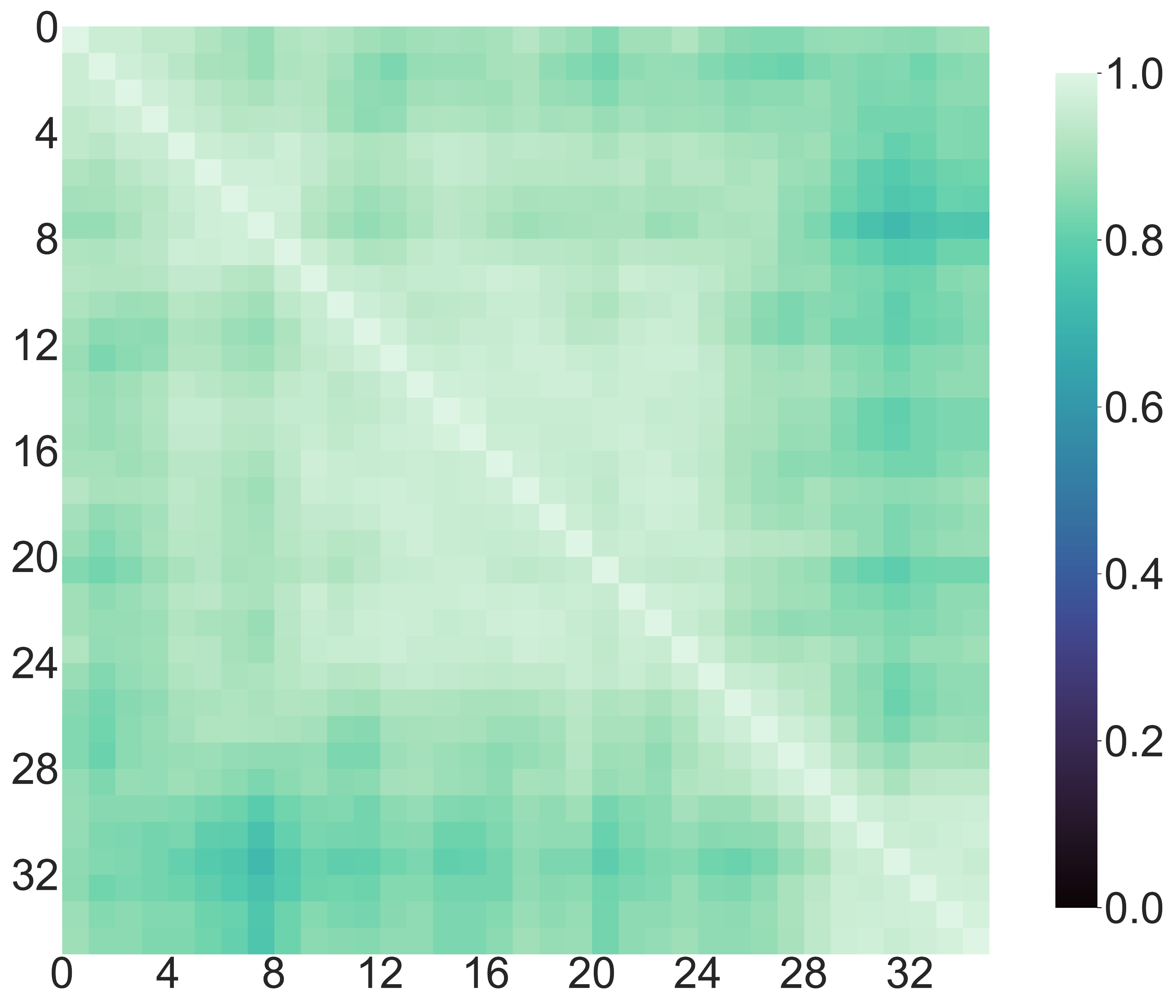}
        \caption{Task Kitchen.}
        
    \end{subfigure}
    \hfill 
    \begin{subfigure}[b]{0.23\textwidth}
        \centering
        \includegraphics[width=\linewidth]{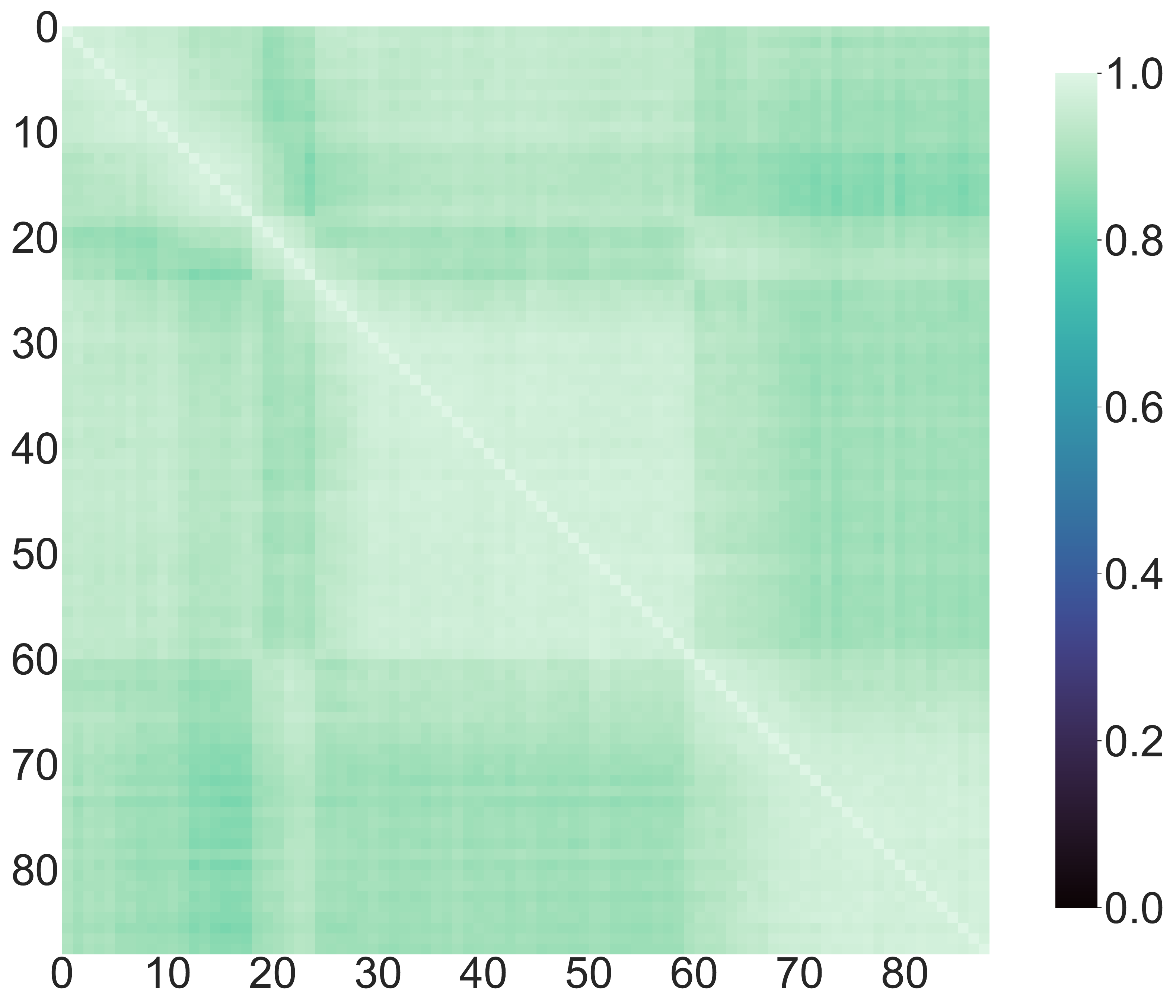}
        \caption{Task Transport.}
        
    \end{subfigure}

    \caption{Similarity between features from different rollouts.}
    \label{fig:similarity}
\end{figure}

\begin{figure}[t]
    \centering 

    \begin{subfigure}[b]{0.23\textwidth}
        \centering
        \includegraphics[width=\linewidth]{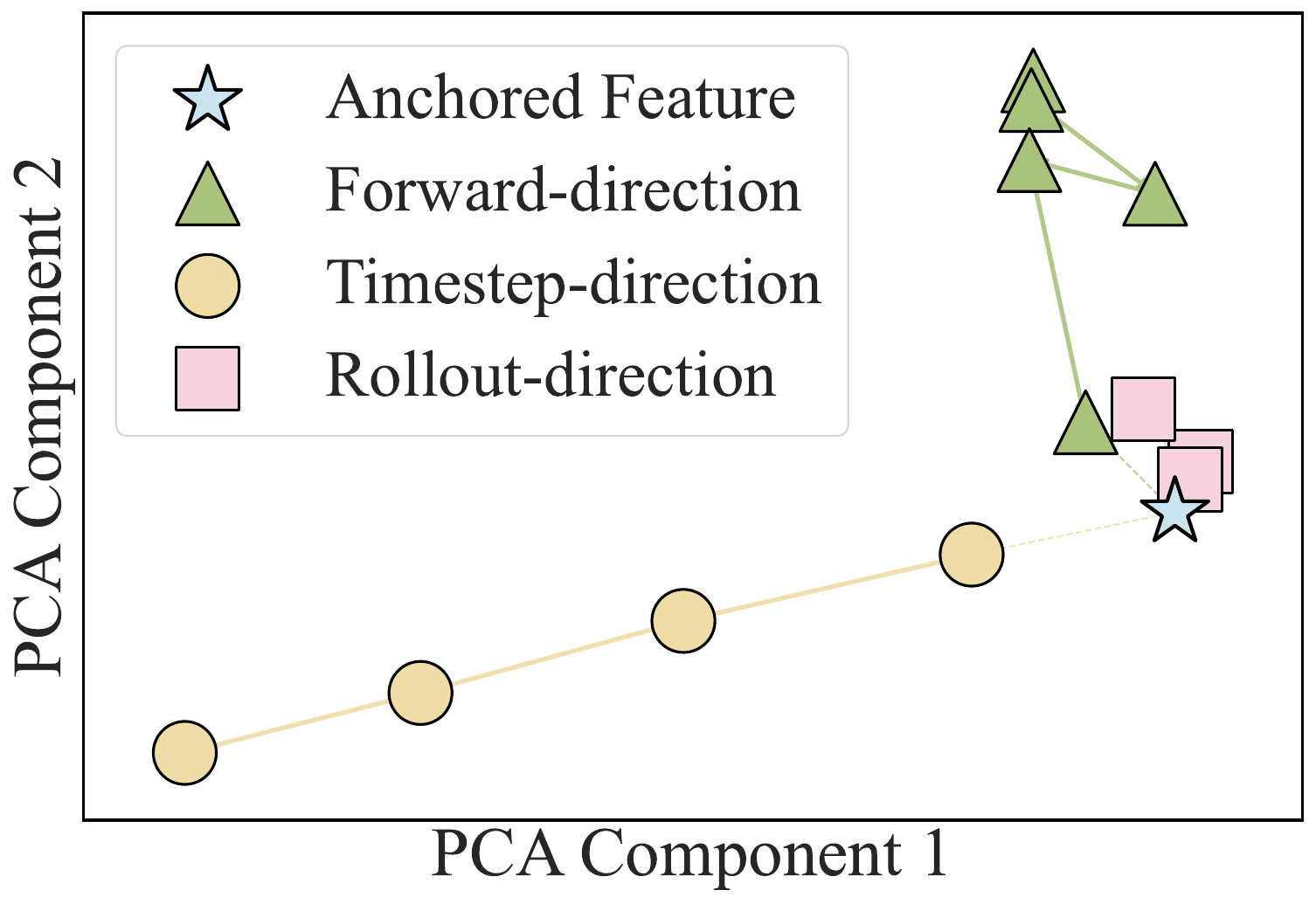}
        \caption{Anchored feature from the 6th self-attention block at the 100th denoising step during the 30th rollout step on the Tool task.}
    \end{subfigure}
    \hfill 
    \begin{subfigure}[b]{0.23\textwidth}
        \centering
        \includegraphics[width=\linewidth]{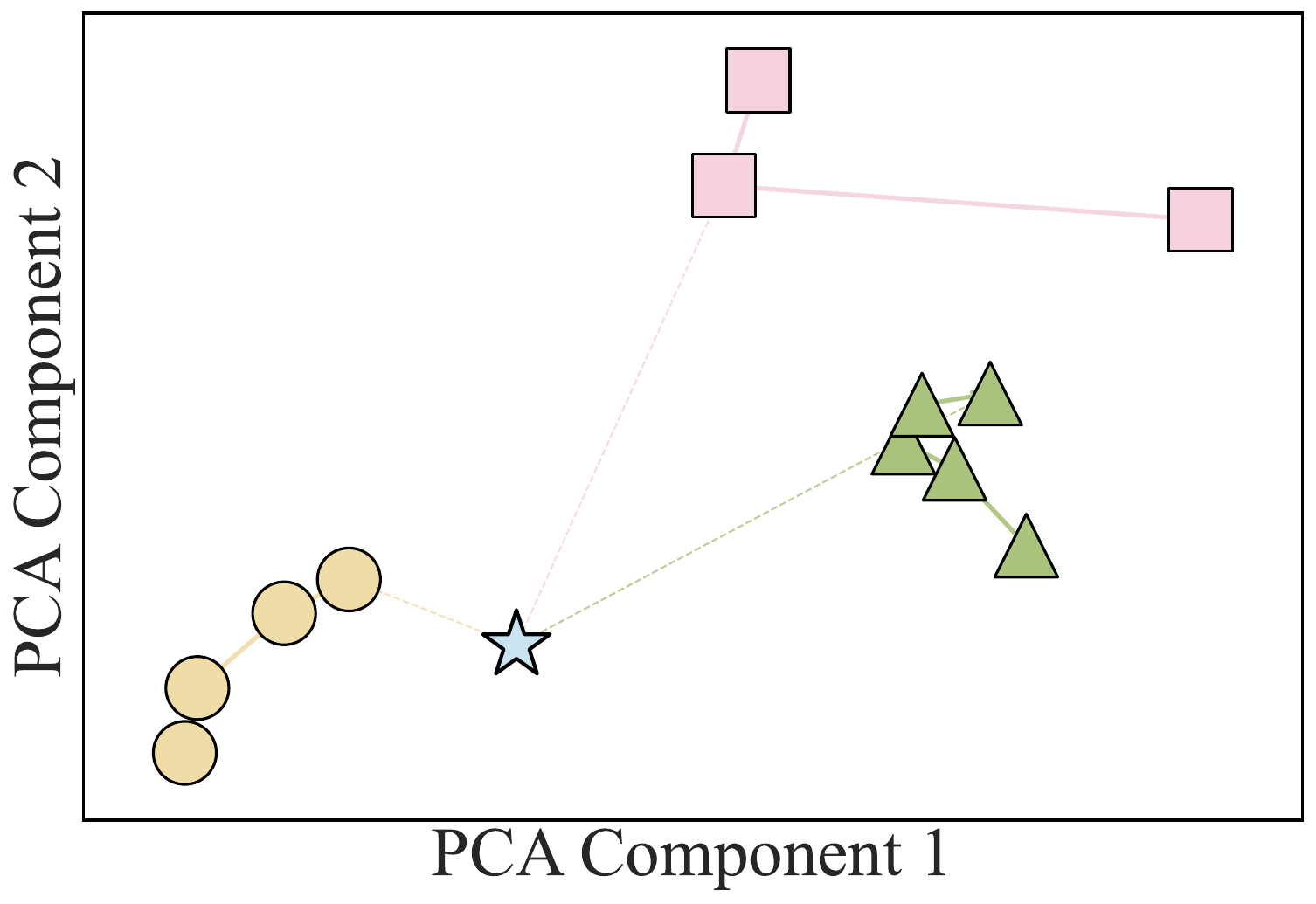}
        \caption{Anchored feature from the 4th self-attention block at the 95th denoising step during the 30th rollout step on the Tool task.}
    \end{subfigure}

    \caption{PCA visualization of features cached from different directions, i.e., the current forward, previous denoising timesteps, and earlier rollout iterations, which are closely aligned with the anchored ones, and offering complementary advantages.}
    \label{fig:PCA}
\end{figure}

While prior caching techniques reuse features from previous timesteps~\cite{ma2024deepcache, ji2025block, yang2025efficientvla}, this single-direction paradigm yields sub-optimal performance under high sparsity (See Table~\ref{tab: ablation}).
Critically, we first observe that features from different rollout iterations exhibit strong similarities, as shown in Figure~\ref{fig:similarity}. Furthermore, by anchoring one feature and visualizing the cached features from various rollout directions, we find that these features are well aligned with the anchored one, each contributing complementary advantages in approaching angles and latent distances, as illustrated in Figure~\ref{fig:PCA}. These observations uncover a significant yet previously underexplored potential for feature reuse.
Motivated by this, we envision an efficient reusing strategy that efficiently utilizes the massive historical features, thereby presenting two critical challenges: 
1) How to efficiently organize the massive features during action diffusion (e.g., ~50k-200k for Diffusion Policy~\cite{chi2023diffusion})? 
2) How to select the most suitable one from them for reuse?



\begin{table*}[t]
  \centering
  \small
  \setlength{\tabcolsep}{3pt}
  \caption{
    \textbf{Benchmark on Proficient Human (PH) demonstration data.} Our method achieves 5.18$\times$ speedup with lossless performance under 95\% sparsity. \colorbox{best!25}{Best} and \colorbox{second!25}{second best} performances are highlighted.
    }
   
  \label{tab:main result 1}
  \begin{tabularx}{\textwidth}{
  l 
  >{\centering\arraybackslash}X 
  *{5}{>{\centering\arraybackslash}X} 
  >{\centering\arraybackslash}X
  >{\centering\arraybackslash}X 
  }
    \toprule
    \multirow{2}{*}{\textbf{Method}} &
    \multirow{2}{*}{\textbf{Sparsity(\%)}} &
    \multicolumn{5}{c}{\textbf{Success Rate (\%, $\uparrow$)} / \textbf{Speedup ($\uparrow$)}} &
    \multirow{2}{*}{\textbf{Average}} &
    \multirow{2}{*}{\textbf{GFLOPS($\downarrow$)}} \\
    \cmidrule(lr){3-7}
     & &  Lift & Can & Square & Transport & Tool & & \\
    \midrule
    \textit{Dense} & 0 & 100 & 94 & 90 & 80 & 50 & 83 & 7.88 \\
    \midrule
    EfficientVLA & 86 & 100 (3.49$\times$) & 74 (3.39$\times$) & 90 (3.53$\times$) & 60 (3.29$\times$) & 38 (3.57$\times$)  & 72 (3.46$\times$) & 1.24  \\
    L2C & 26 &  100 (1.48$\times$) & 86 (1.27$\times$) & 26 (1.31$\times$) & 66 (1.32$\times$) & 2 (1.33$\times$)  & 56 (1.28$\times$) & 5.87 \\ 
    BAC & 90 & 100 (3.66$\times$) & \cellcolor{second!25}{94 (3.78$\times$)} & \cellcolor{best!25}{94 (3.79$\times$)} & 84 (3.43$\times$) & 26 (3.74$\times$)  & 79 (3.68$\times$) & 1.07 \\    
    \midrule 
    Ours & \cellcolor{second!25}{93} &\cellcolor{second!25}{100 (4.90$\times$)} & \cellcolor{best!25}{94 (4.96$\times$)} & {90} ({4.85}$\times$) & \cellcolor{second!25}{92 (4.70$\times$)} & \cellcolor{best!25}{56 (4.88$\times$)}  & \cellcolor{best!25}{86 (4.86$\times$)} & \cellcolor{second!25}{0.68}  \\
   Ours & \cellcolor{best!25}{95} & \cellcolor{best!25}{ 100 (5.31$\times$)} & {88} ({5.39}$\times$) & \cellcolor{second!25}{92 (5.18$\times$)} & \cellcolor{best!25}{94 (4.83$\times$)} & \cellcolor{second!25}{48 (5.21$\times$)}  & \cellcolor{second!25}{84 (5.18$\times$)} & \cellcolor{best!25}{0.42} \\
    \bottomrule
  \end{tabularx}
\end{table*}


\paragraph{3D Lattice Modeling.}

To address the first challenge, we first model the historical feature space as a 3D lattice defined by three orthogonal indices: block index $b$, denoising timestep $k$, and rollout iteration $r$. Each anchored feature exists at a coordinate $(b, k, r)$ in this space.
Our omnidirectional reusing strategy organizes the cache by maintaining the closest available feature along each of these three axes. 
As illustrated in Figure~\ref{fig:overview}, a cached feature differs from its anchored counterpart by only one index.
For example, a feature cached from a previous rollout has coordinates $(b, k, r-1)$, sharing the same block and timestep indices. Globally, we maintain separate, dedicated buffers for all three directions, retaining only the most recently updated feature along each axis. This structure efficiently assigns three candidates for each anchored feature.

\paragraph{Trajectory-level Training.}
To address the second challenge, we incorporate the cache-selection mechanism directly into the pruner. For the $b$-th block at denosing timestep $k$, the pruner outputs a 4-dimensional gating vector
\begin{equation}
\mathbf{p}_{b,k} = (p^{C}_{b,k}, p^{F}_{b,k}, p^{T}_{b,k}, p^{R}_{b,k}),
\end{equation}
where each element represents the confidence in either performing a new computation or reusing a feature cached from the forward, denoising timestep, or rollout directions. During inference, the gating vector is discretized via an $\arg\max$ operation to yield the mask vector $\mathcal{M}_{b,k}$. To enable gradient-based optimization, we adopt the Straight-Through Estimator~\cite{jang2016ste} during training, which allows the gradient to flow through the non-differentiable $\arg\max$ operation while maintaining discrete gating behavior in the forward pass.
The residual update is thus formulated as

\begin{equation}
\begin{aligned}
h_{k}^{\left\lfloor b/3 \right\rfloor}
= \; & h_{k}^{\left\lfloor b/3 \right\rfloor-1}
+ \mathcal{M}^{C}_{b,k} d_{b,k}
+ \mathcal{M}^{F}_{b,k}\tau^{F}_{b} \\
& + \mathcal{M}^{T}_{b,k}\tau^{T}_{b}
+ \mathcal{M}^{R}_{b,k}\tau^{R}_{b,k}.
\end{aligned}
\end{equation}
where $d_{b,k}$ denotes the newly computed feature, and $\tau^{F}_{b}$, $\tau^{T}_{b}$, and $\tau^{R}_{b,k}$ are features cached from different directions. 
  The corresponding caches are refreshed according to:
\begin{align}
\tau^{F}_{b} &\leftarrow (1 - {\mathcal{M}}^{C}_{b,k})\tau^{F}_{b} + {\mathcal{M}}^{C}_{b,k} d_{b,k} \\
\tau^{T}_{b} &\leftarrow (1 - {\mathcal{M}}^{C}_{b,k})\tau^{T}_{b} + {\mathcal{M}}^{C}_{b,k}d_{b,k} \\
\tau^{R}_{b,k} &\leftarrow (1 - {\mathcal{M}}^{C}_{b,k})\tau^{R}_{b,k} + {\mathcal{M}}^{C}_{b,k} d_{b,k}
\end{align}

To learn the reuse strategy along the rollout direction, we sample a few action trajectories and supervise the sparsified diffusion outputs step by step across the rollout iterations. 
During each trajectory, the pruner predicts a binary mask $\mathcal{M}^{r}$ in each rollout iteration $r$, and the action diffusion sparsified by $\mathcal{M}^{r}$ outputs an action $\hat{a}_{r}$. 
After each diffusion step, gradients are backpropagated, thus enabling multi-step supervision throughout the entire trajectory. 
In each iteration, the pruner is jointly optimized with a fidelity loss $\mathcal{L}_{\mathrm{f}}$ and a sparsity regularization loss $\mathcal{L}_{\mathrm{s}}$:
\begin{equation}
\mathcal{L} = \mathcal{L}_{\mathrm{f}} + \mathcal{L}_{\mathrm{s}},
\end{equation}
where
\begin{align}
\mathcal{L}_{\mathrm{f}} &= 
\mathbb{E}_{(o_r,a^*_r)\sim\mathcal{D}_{\text{ref}}}
\big[\|\pi'(\mathbf{o_r},\mathcal{M}^r)-a^*_r\|\big],\\
\mathcal{L}_{\mathrm{s}} &= \left\lvert
\frac{1}{BK}\sum_{b=1}^{B}\sum_{k=1}^{K}p^{c}_{b,k}
- (1-\rho)
\right\rvert.
\end{align}
Here, $a^{*}_{r}$ is the reference action at rollout iteraion $r$, and $\rho$ represents the target pruning rate.









\section{Experiment}
We first outline the experimental setup in Section~\ref{sec: exp setup}. Following that, we demonstrate the quantitative results in Section~\ref{sec: main result}. We conduct an ablation study of the method in Section~\ref{sec: ablation study}, and present qualitative analysis in Section~\ref{sec: qualitative results} .

\subsection{Experimental Setup}
\label{sec: exp setup}

\begin{table*}[t]
  \centering
  \small
  \setlength{\tabcolsep}{3pt}
  \caption{
      \textbf{Benchmark on Mixed Human (MH) demonstration data.}
      Our method achieves over 4.6$\times$ speedup across all tasks. \colorbox{best!25}{Best} and \colorbox{second!25}{second best} performances are highlighted.
  }
  \label{tab:main result 2}
  \begin{tabularx}{\textwidth}{l >{\centering\arraybackslash}X *{4}{>{\centering\arraybackslash}X} >{\centering\arraybackslash}X >{\centering\arraybackslash}X} 
    \toprule
    \multirow{2}{*}{\textbf{Method}} &
    \multirow{2}{*}{\textbf{Sparsity(\%)}} &
    \multicolumn{4}{c}{\textbf{Success Rate (\%, $\uparrow$)} / \textbf{Speedup ($\uparrow$)}} &
    \multirow{2}{*}{\textbf{Average}} &
    \multirow{2}{*}{\textbf{GFLOPS($\downarrow$)}}\\
    \cmidrule(lr){3-6}
    & & Lift & Can & Square & Transport & & \\
    \midrule
    \textit{Dense} & 0 & 100 & 94 & 74 & 56 & 81 & 7.88 \\
    \midrule
    EfficientVLA & 86 & 100 (3.45$\times$) & 76 (3.31$\times$) & 52 (3.37$\times$) & 0 (3.69$\times$) & 57 (3.46$\times$) & 1.24 \\
    L2C &26 & 100 (1.28$\times$) & 0 (1.33$\times$) & 54 (1.25$\times$) & 46 (1.33$\times$) & 50 (1.30$\times$) & 5.87 \\ 
    BAC & 90 & 100 (3.69$\times$) & 70 (3.73$\times$) & \cellcolor{second!25}{78 (3.72$\times$)} & 50 (3.65$\times$) & 74 (3.70$\times$) & 1.06 \\
    \midrule
    \textbf{Ours} & \cellcolor{second!25}{93} & \cellcolor{second!25}{100 (4.86$\times$)} & \cellcolor{second!25}{94 (4.94$\times$)} & \cellcolor{best!25}{82 (4.83$\times$)} & \cellcolor{best!25}{62 (4.69$\times$)} & \cellcolor{best!25}{85 (4.83$\times$)} & \cellcolor{second!25}{0.72}  \\
    \textbf{Ours} & \cellcolor{best!25}{95} & \cellcolor{best!25}{100 (5.30$\times$)} & \cellcolor{best!25}{94 (5.30$\times$)} & 74 (5.24$\times$) & \cellcolor{second!25}{50 (5.02$\times$)} & \cellcolor{second!25}{80 (5.22$\times$)} & \cellcolor{best!25}{0.44} \\

    \bottomrule
  \end{tabularx}
\end{table*}

\begin{table*}[t]
  \centering
  \small
  \setlength{\tabcolsep}{3pt}
  \caption{
      \textbf{Benchmark on multi-stage task (Kitchen).} 
      In the Kitchen task, $p_x$ is the frequency of interacting with $x$ or more objects, including an openable microwave, four turnable oven burners, an oven light switch, and a freely movable kettle.
      Our method achieves a lossless 5.90$\times$ speedup on this task. \colorbox{best!25}{Best} and \colorbox{second!25}{second best} performances are highlighted.
  }
  \label{tab:main result 3}
  \begin{tabularx}{\textwidth}{l >{\centering\arraybackslash}X *{6}{>{\centering\arraybackslash}X}>{\centering\arraybackslash}X}
    \toprule
    \multirow{2}{*}{\textbf{Method}} &
    \multirow{2}{*}{\textbf{Sparsity(\%)}} & 
    \multicolumn{5}{c}{\textbf{Success Rate (\%, $\uparrow$)}} &
    \multirow{2}{*}{\textbf{Speedup} ($\uparrow$)} &
    \multirow{2}{*}{\textbf{GFLOPS}($\downarrow$)}\\
    \cmidrule(lr){3-7}
     & & Kit$_{p1}$ & Kit$_{p2}$ & Kit$_{p3}$ & Kit$_{p4}$ & Average &\\
    \midrule
    \textit{Dense} & 0 & 100 & 100 & 100 & 100 & 100 & -- & 113 \\
    \midrule
    EfficientVLA & 86 & 20 & 2 & 0 & 0 & 3 & 3.60$\times$&13.81\\
    L2C & 25 & 100 & 100 & \cellcolor{second!25}{100} & 98 & 99 & 1.28$\times$ & 85.5 \\  
    BAC & 90 & 100 & 98 & 94 & 82 & 93 & 3.90$\times$ & 15.83 \\
    \midrule
    \textbf{Ours} & \cellcolor{second!25}{93} &  \cellcolor{second!25}{100} & \cellcolor{second!25}{100} & \cellcolor{best!25}{100} & \cellcolor{best!25}{100} & \cellcolor{best!25}{100} & \cellcolor{second!25}{5.90$\times$} & \cellcolor{second!25}{9.71} \\ 
    \textbf{Ours} & \cellcolor{best!25}{95} & \cellcolor{best!25}{100} & \cellcolor{best!25}{100} & 98 & \cellcolor{second!25}{98} & \cellcolor{second!25}{99}& \cellcolor{best!25}{6.33$\times$} & \cellcolor{best!25}{7.28} \\ 
    \bottomrule
  \end{tabularx}
\end{table*}

\noindent \textbf{Models and Benchmarks.} 
We evaluate the acceleration performance of our method on two types of action diffusion models: diffusion policy and vision-language-action (VLA) model, using the most representative architectures from each category.
For diffusion policy, we adopt the transformer-based Diffusion Policy~\cite{chi2023diffusion} as our primary evaluation backbone. Our method is tested across a range of robot manipulation tasks with fixed seeds, including Lift, Can, Square, Transport, Tool Hang, and Kitchen~\citep{gupta2020relay}. The demonstration datasets consist of trajectories collected from proficient human (PH) and mixed proficient/non-proficient human (MH) teleoperation.
To further assess the generality of our approach under VLA settings, we extend our experiments to the state-of-the-art RDT-1B model~\cite{liu2024rdt}, which incorporates a 1-billion-parameter diffusion transformer for action generation, coupled with T5-XXL for language conditioning and SigLIP for visual encoding. We evaluate our method on four ManiSkill simulation tasks: PushCube, PickCube, StackCube, and Insert.

 \noindent \textbf{Metrics.} We assess the performance of different methods using the \textit{Success Rate} over 50 episodes, following prior works~\citep{chi2023diffusion}. We evaluate the efficiency by reporting \textit{Sparsity}, \textit{Speedup}, and \textit{GFLOPS}. \textit{Sparsity} indicates the percentage of pruned computations during inference. \textit{Speedup} measures the wall-clock time reduction compared to the full-precision model. \textit{GFLOPS} quantifies the total floating-point operations required for each action diffusion.

\noindent \textbf{Baselines.} 
We establish the original models as our \textit{Dense} baseline. For comparison, we benchmark against leading acceleration methods. We reproduce current cache-based acceleration methods for action generation, Efficient-VLA~\citep{yang2025efficientvla} and BAC~\citep{ji2025block}. Additionally, we adapt L2C~\citep{ma2024learningtocache}, a learning-based acceleration method originally designed for image generation, for our action generation tasks.

\noindent \textbf{Implementation Details.} 
 The experiments are conducted on NVIDIA Tesla A40 48G GPU, equipped with Intel(R) Xeon(R) Gold 6230 CPU@2.10GHz. The pruner introduces a hyperparameter pruning rate $\rho$, which we set as 80\%,90\%, 93\%, and 95\% for different settings. The pruner network is trained for $20$ epochs with a learning rate of $1e^{-4}$ and weight decay of $1e^ {-4}$. We set the batch size as 16 and 1 for Diffusion Policy and RDT-1B, respectively.



\subsection{Quantitative Results}
\label{sec: main result}
For Diffusion Policy, we test two sampling configurations: DDPM (100 denoising steps) and DDIM (40 denoising steps). For RDT-1B, we follow the original paper and adopt DPM-Solver with 50 steps. For L2C, we adopt the same training configs as our method for fair comparison. We set the cache updating interval $\mathcal N$ as 7 for EfficientVLA. We set the number of cache update steps $\mathcal S$ as 8 and the number of selected upstream blocks $k$ as 5 for BAC. 
We conduct quantitative evaluations as shown in Tables~\ref{tab:main result 1}, \ref{tab:main result 2}, \ref{tab:main result 3}, \ref{tab:DDIM} and \ref{tab:RDT-1B}.


\noindent \textbf{Effectiveness on Diffusion Policy.} 
As shown in Tables~\ref{tab:main result 1}, our method prunes over 95\% computations during action diffusion while achieving lossless performance on PH dataset, demonstrating great sparsity of action diffusion. Moreover, Tables~\ref{tab:main result 1}, ~\ref{tab:main result 2}, and ~\ref{tab:main result 3} report that our method achieves comparable performance with the original model under 93\% sparsity across all tasks, reaching up to 6.33$\times$ speedup on the challenging multi-stage Kitchen task.




\noindent \textbf{Collaboration with Efficient Sampling Methods.}
As shown in Tables~\ref{tab:DDIM} and~\ref{tab:RDT-1B}, our method also demonstrates effectiveness when combined with efficient samplers like DDIM and DPM-Solver. On the Diffusion Policy with DDIM sampler using 40 denoising steps, our method achieves a lossless performance with over 3$\times$ speedup under 80\% sparsity across all tasks. For RDT-1B with DPM-Solver using 50 denoising steps, our method attains over 2.5$\times$ speedup under 90\% sparsity while maintaining or even improving the success rates on all ManiSkill tasks.

\begin{table}[!htbp]
    \centering
    \small
    \setlength{\tabcolsep}{3pt}
    \caption{\textbf{Acceleration on DDIM}. We evaluate Diffusion Policy with a DDIM sampler using 40 denoising steps on MH datasets under 80\% sparsity. \colorbox{best!25}{Best} performances are highlighted.}
    \label{tab:DDIM}

  \vspace{0.5 em}
      \begin{tabularx}{\linewidth}{l *{4}{>{\centering\arraybackslash}X}}
      \toprule
      \multirow{2}{*}{\textbf{Method}} &
      \multicolumn{4}{c}{  \textbf{Success Rate (\%, $\uparrow$)} / \textbf{Speedup ($\uparrow$)}} \\
      \cmidrule(lr){2-5}
      &  Kitchen &Lift  &  Square &  Transport \\
      \midrule
      \textit{Dense}  & 98 & 100 & 78  & 52  \\ 
      \midrule
      
      \textbf{Ours}
      & \cellcolor{best!25}{100} (3.64$\times$) & \cellcolor{best!25}{100} (2.96$\times$) & \cellcolor{best!25}{82} (3.06$\times$) & \cellcolor{best!25}{56} (2.87$\times$) \\

  \bottomrule
  \end{tabularx}
\end{table}

\noindent \textbf{Effectivenss on VLA.} 
We report the acceleration results on the large-scale RDT-1B model in Table~\ref{tab:RDT-1B}. Under 90\% sparsity, our method achieves significant speedups of over 2.5$\times$ across all ManiSkill tasks while maintaining or even improving the success rates compared to the original model. We observe the lower speedup compared to Diffusion Policy, mainly due to the heavier visual and language encoders in RDT-1B, which limit the end-to-end acceleration.


\begin{table}[!htbp]
    \centering
    \small
    \setlength{\tabcolsep}{3pt}
    \caption{\textbf{Quantitative Results on VLA.} We evaluate our approach on the large-scale RDT-1B model across ManiSkill manipulation tasks under 90\% sparsity. \colorbox{best!25}{Best} results are highlighted.
}
    \label{tab:RDT-1B}
  \vspace{0.5 em}
      \begin{tabularx}{\linewidth}{l *{4}{>{\centering\arraybackslash}X}}
      \toprule
      \multirow{2}{*}{\textbf{Method}} &
      \multicolumn{4}{c}{\textbf{Success Rate (\%, $\uparrow$)} / \textbf{Speedup ($\uparrow$)}} \\
      \cmidrule(lr){2-5}
      &  Push & Pick &  Stack &  Insert \\
      \midrule
      \textit{Dense}  & 100 & 76 & 80  & 4  \\ 
      \midrule
      
      \textbf{Ours} 
      & \cellcolor{best!25}{100} (3.11$\times$) & \cellcolor{best!25}{80} (2.64$\times$) & \cellcolor{best!25}{80} (2.49$\times$) & \cellcolor{best!25}{16} (2.67$\times$) \\

  \bottomrule
  \end{tabularx}
\end{table}

\noindent \textbf{Comparison with SOTA baselines.} 
Our method shows significant advantages over existing state-of-the-art acceleration methods, which fail to deliver comparable performance under high sparsity. L2C~\citep{ma2024learningtocache} forces computing the features before each reuse step, resulting in a modest 1.28$\times$ speedup. EfficientVLA follows a naive reusing strategy that reuses in a fixed interval, which leads to poor performance, such as a 3\% success rate on the Kitchen task. Despite BAC achieving fine-grained block-wise caching, it cannot adapt to the evolving rollout dynamics, leading to a sub-optimal trade-off between performance and efficiency.
These baseline methods are limited to reusing features cached solely along the denoising-step dimension, which significantly restricts the achievable speedup. 
In contrast, our approach constrains the prune-induced errors successfully by the omnidirectional reusing strategy that jointly reuses features from massive historical features, which preserves the high fidelity of the generated action under even 95\% sparsity. The method further transfers the advantage into 5$\times$ practical speedup by the parallelized inference pipeline.

\subsection{Ablation Study}
\label{sec: ablation study}

To verify the effectiveness of Omnidirectional Reusing, we conduct an ablation study by reusing the features from only a single specific direction. We evaluate the variants on the PH dataset under 93\% sparsity. As summarized in Table~\ref{tab: ablation}. all the single-direction variants suffer noticeable drops in success rate across all tasks. Moreover, each variant encounters severe performance degradation on certain tasks, such as Forward-direction on Transport, Timestep-direction on Tool, and Rollout-direction on Can, indicating that reusing features along only one direction is insufficient to compensate for the aggressive pruning. In contrast, the full omnidirectional caching scheme consistently achieves the best trade-off between performance and efficiency by selectively reusing cached features from all directions.


\begin{table}[!htbp]
  \centering
  \small
  \setlength{\tabcolsep}{3pt} 
  \caption{
    \textbf{Ablation Study.}
    We report the success rate when only reusing features from single directions on PH dataset under 93\% sparsity. \colorbox{best!25}{Best} performances are highlighted.
  }
  \label{tab: ablation}
  
  \vspace{0.5 em}
      \begin{tabularx}{\linewidth}{l *{4}{>{\centering\arraybackslash}X}}
      \toprule
      \multirow{2}{*}{\textbf{Method}} &
      \multicolumn{4}{c}{\textbf{Success Rate (\%, $\uparrow$)}} \\
      \cmidrule(lr){2-5}
      & Can & Transport & Tool & Square \\
      \midrule
      \textit{Dense}    & 94 & 80 & 50  & 90 \\
      \midrule
      Forward-direction  & 86 & 4 & 50  & 18  \\
      Timestep-direction & 86 & 78  & 0 & 80 \\
      Rollout-direction & 10 & 70 & 32 & 80 \\    
      \midrule
      \textbf{Omini-direction} & \cellcolor{best!25}{94} & \cellcolor{best!25}{92} & \cellcolor{best!25}{56} & \cellcolor{best!25}{90} \\
  \bottomrule
  \end{tabularx}
\end{table}

\subsection{Qualitative Analysis}
\label{sec: qualitative results}
\begin{figure}[htbp]
    \centering 
    
    \begin{subfigure}[b]{0.5\textwidth} 
        \centering
        \includegraphics[width=\textwidth]{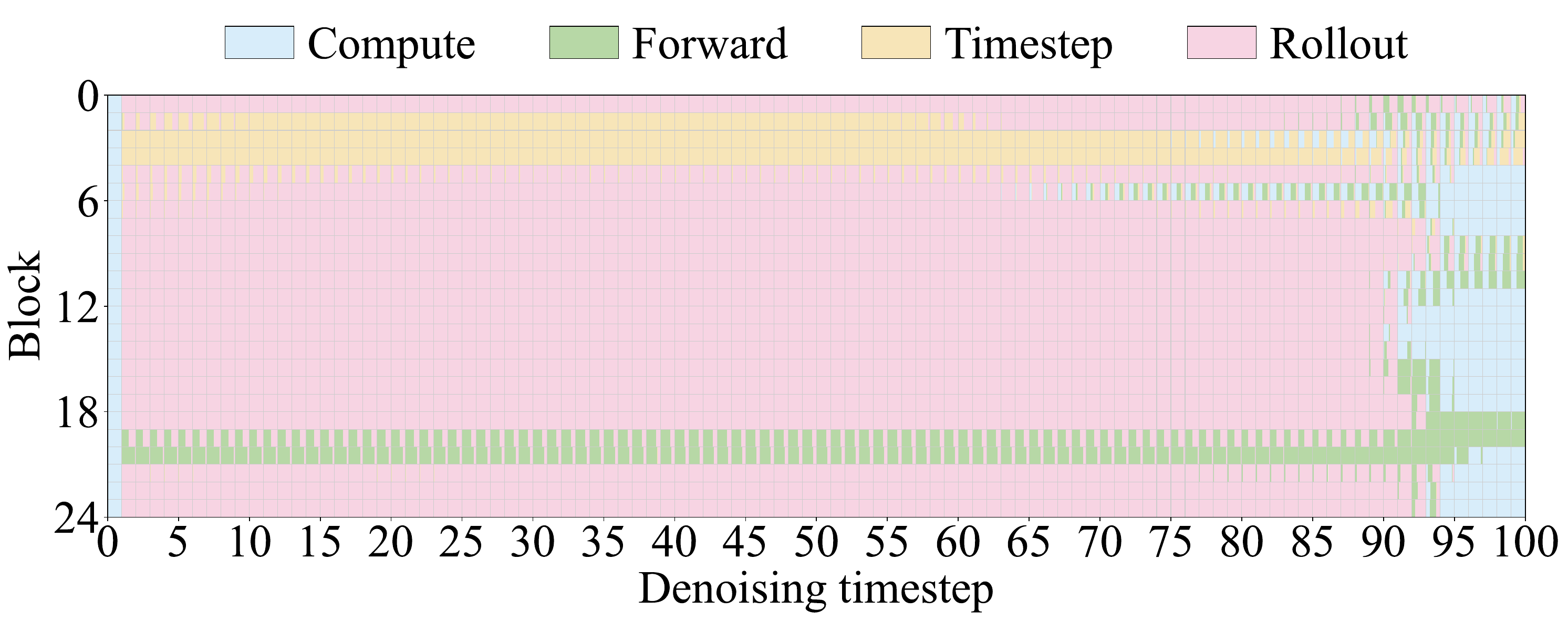}
        \caption{The 1-th Rollout Iteration}
    \end{subfigure}

    \begin{subfigure}[b]{0.5\textwidth}
        \centering
        \includegraphics[width=\textwidth]{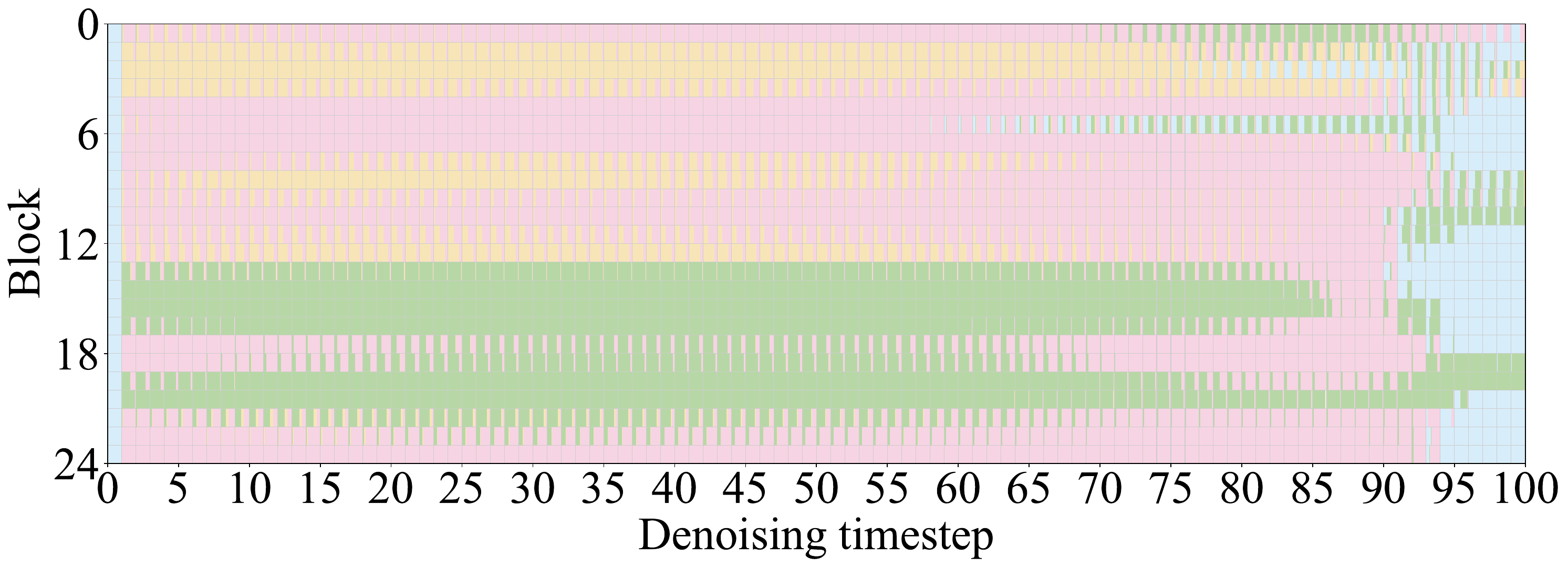}
        \caption{The 20-th Rollout Iteration}
    \end{subfigure}
    
    \begin{subfigure}[b]{0.5\textwidth}
        \centering
        \includegraphics[width=\textwidth]{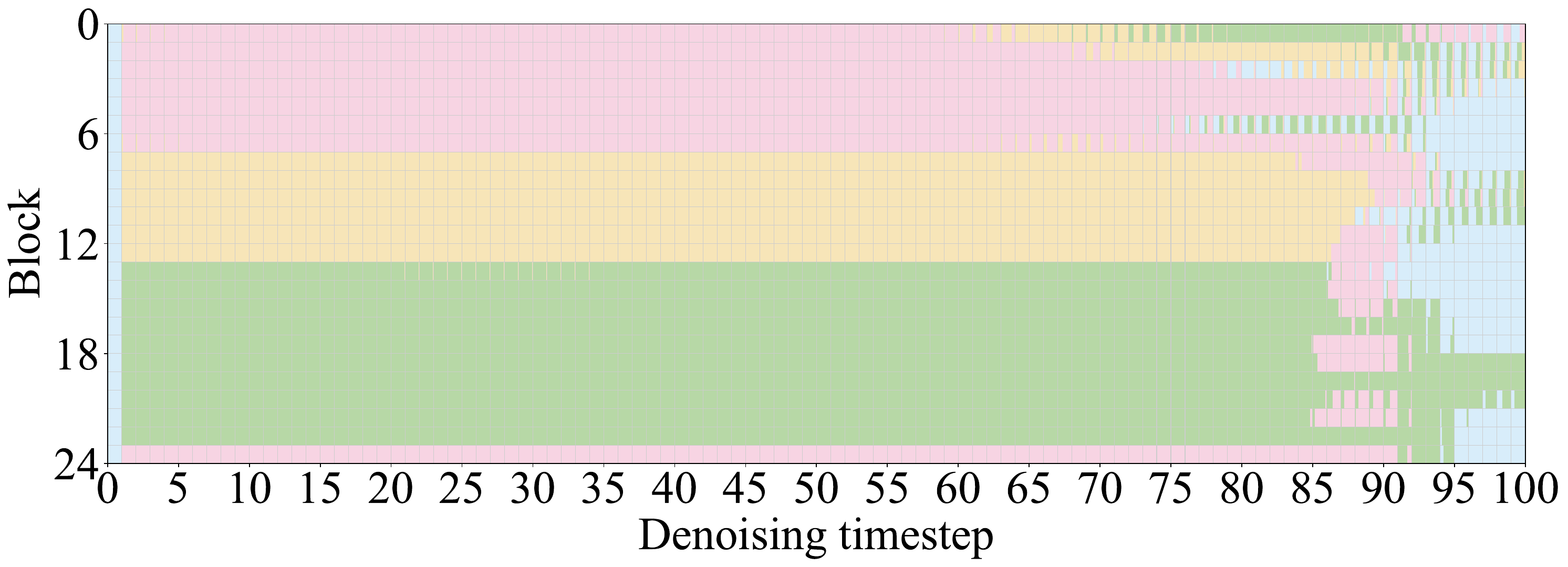}
        \caption{The 40-th Rollout Iteration}
    \end{subfigure}
    
    \caption{\textbf{Visualization of $\mathcal{M}$ during the robot lifting a small box.} Each cell at coordinate $(k,b)$ denotes $\mathcal{M}_{b,k}$. The proportions of the colored regions within each cell indicate the confidence scores $p^{c}_{b,k}$, $p^{F}_{b,k}$, $p^{T}_{b,k}$, and $p^{R}_{b,k}$.}
    \label{fig:cache_main} 
\end{figure}

Figure~\ref{fig:cache_main} visualizes the predicted pruning mask $\mathcal{M}$. The mask $\mathcal{M}$ exhibits significant variations throughout the rollout iterations, revealing the presence of test-time sparsity that emerges alongside the evolving visuomotor dynamics in an open environment. Moreover, each direction occupies considerable proportions, highlighting the necessity of aggregating historical features from all directions. More details are provided in the supplementary material.


\section{Conclusion}
In this work, we propose test-time sparsity, which accelerates the action diffusion by 5$\times$, achieving 47.5Hz inference frequency without performance degradation. The result benefits from two key designs: First, we design a highly parallelized inference pipeline that minimizes the non-decoder delay to milliseconds. Second, we introduce omnidirectional reusing, which achieves 95\% sparsity by selectively reusing the historical features cached from the current forward, previous denoising timesteps, and earlier rollout iterations.

\section{Acknowledgement}
This work is supported by the National Natural Science Foundation of China (Grant No. 92467204, 62472249, and 62402264), and the Shenzhen Science and Technology Program (Grant No. JCYJ20220818101014030 and KJZD20240903102300001).
{
    \small
    \bibliographystyle{ieeenat_fullname}
    \bibliography{reference}
}

\appendix
\clearpage
\maketitlesupplementary

\section{Details on Rollout Similarity}

\begin{figure}[htbp]
    \centering 
    \begin{subfigure}[b]{0.23\textwidth}
        \centering
        \includegraphics[width=\linewidth]{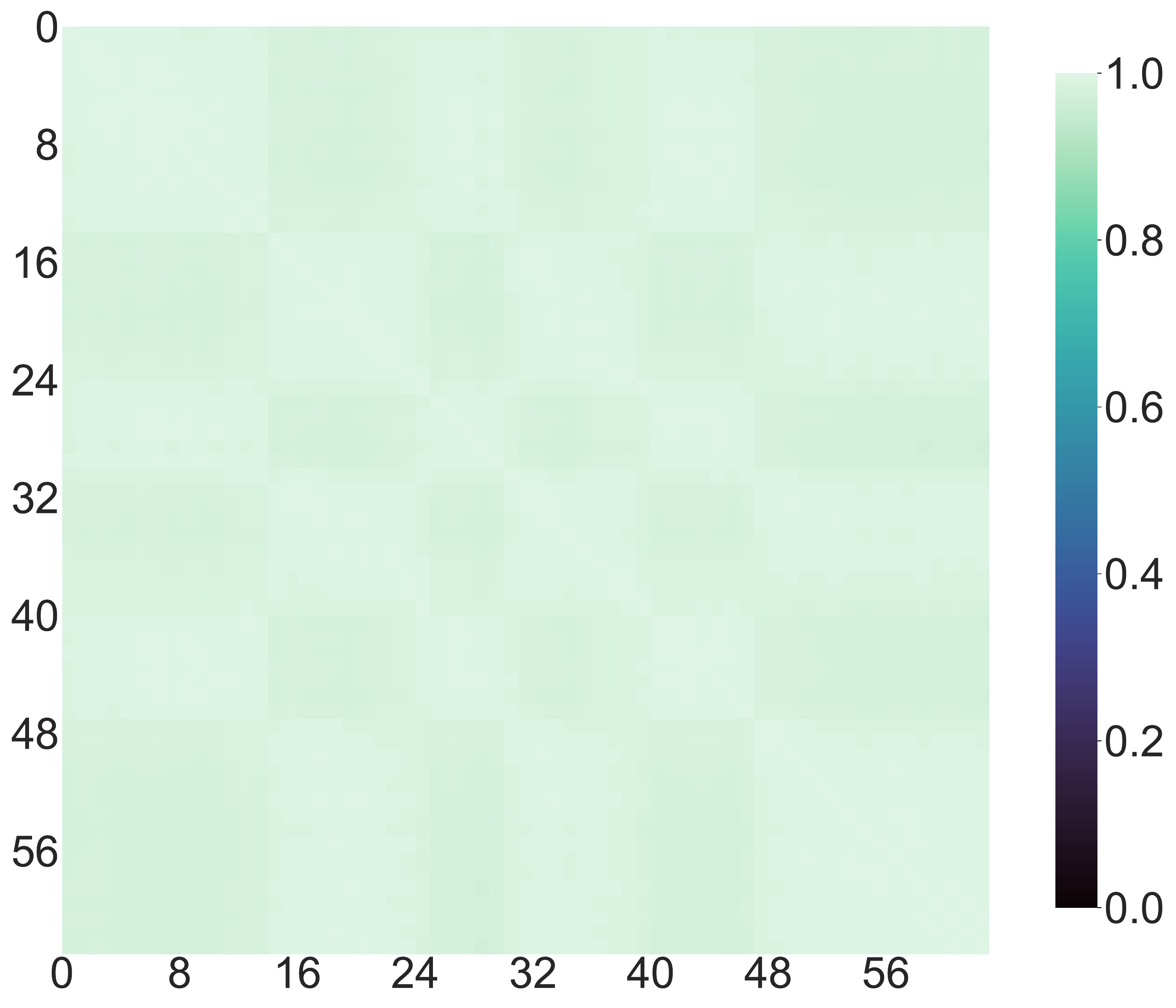}
        \caption{Task Lift.}
        
    \end{subfigure}
    \hfill 
    \begin{subfigure}[b]{0.23\textwidth}
        \centering
        \includegraphics[width=\linewidth]{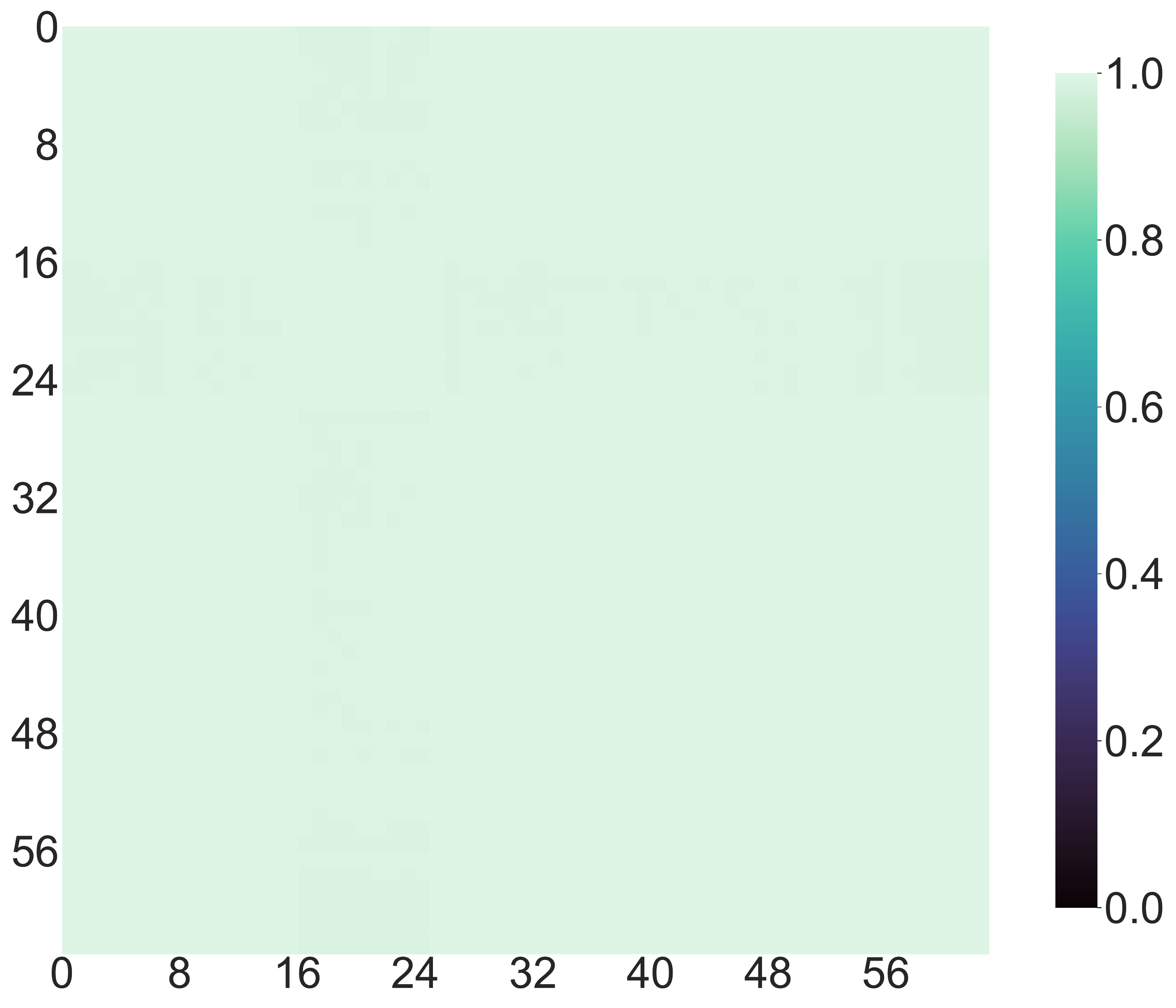}
        \caption{Task Square.}
        
    \end{subfigure}

    \begin{subfigure}[b]{0.23\textwidth}
        \centering
        \includegraphics[width=\linewidth]{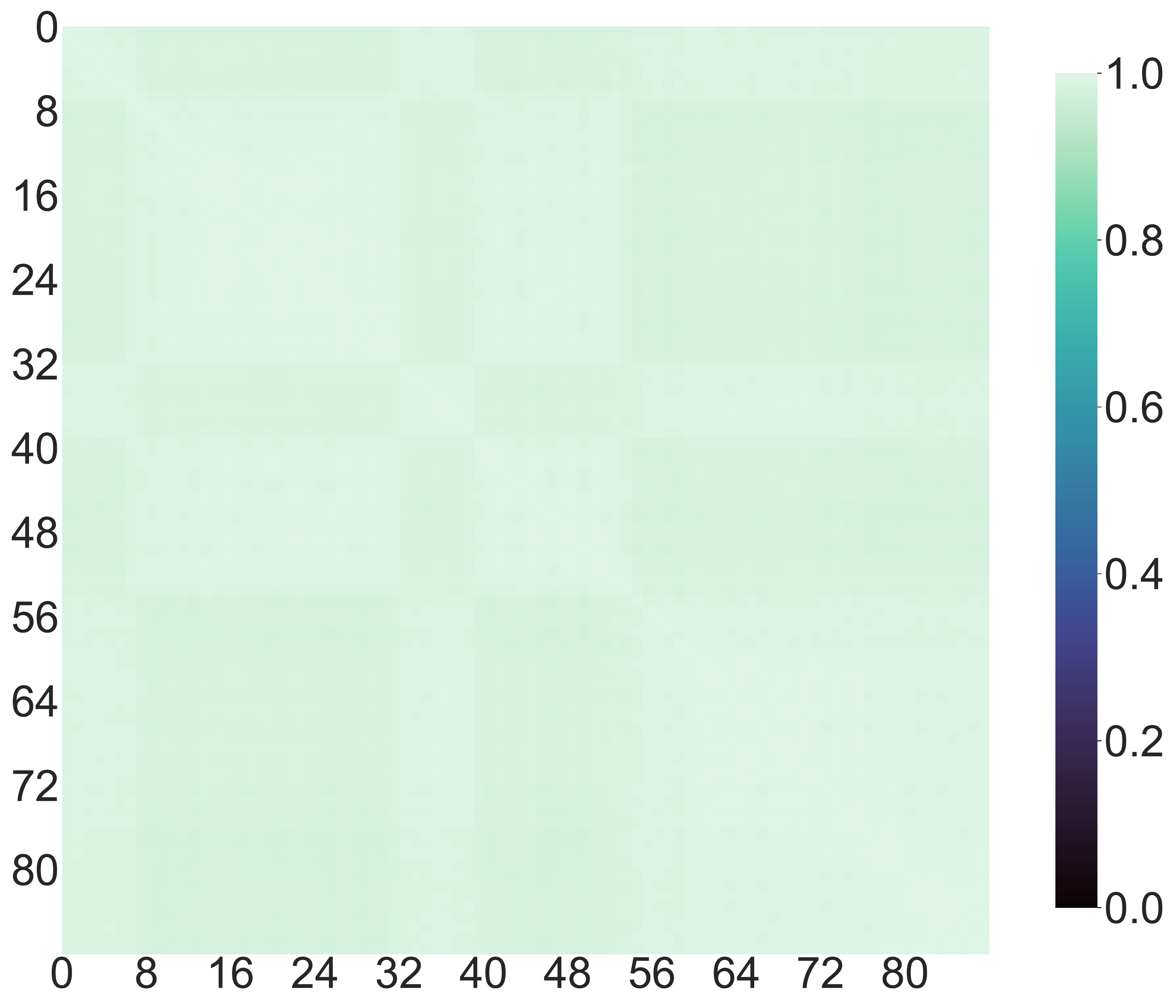}
        \caption{Task Tool.}
        
    \end{subfigure}
    \hfill 
    \begin{subfigure}[b]{0.23\textwidth}
        \centering
        \includegraphics[width=\linewidth]{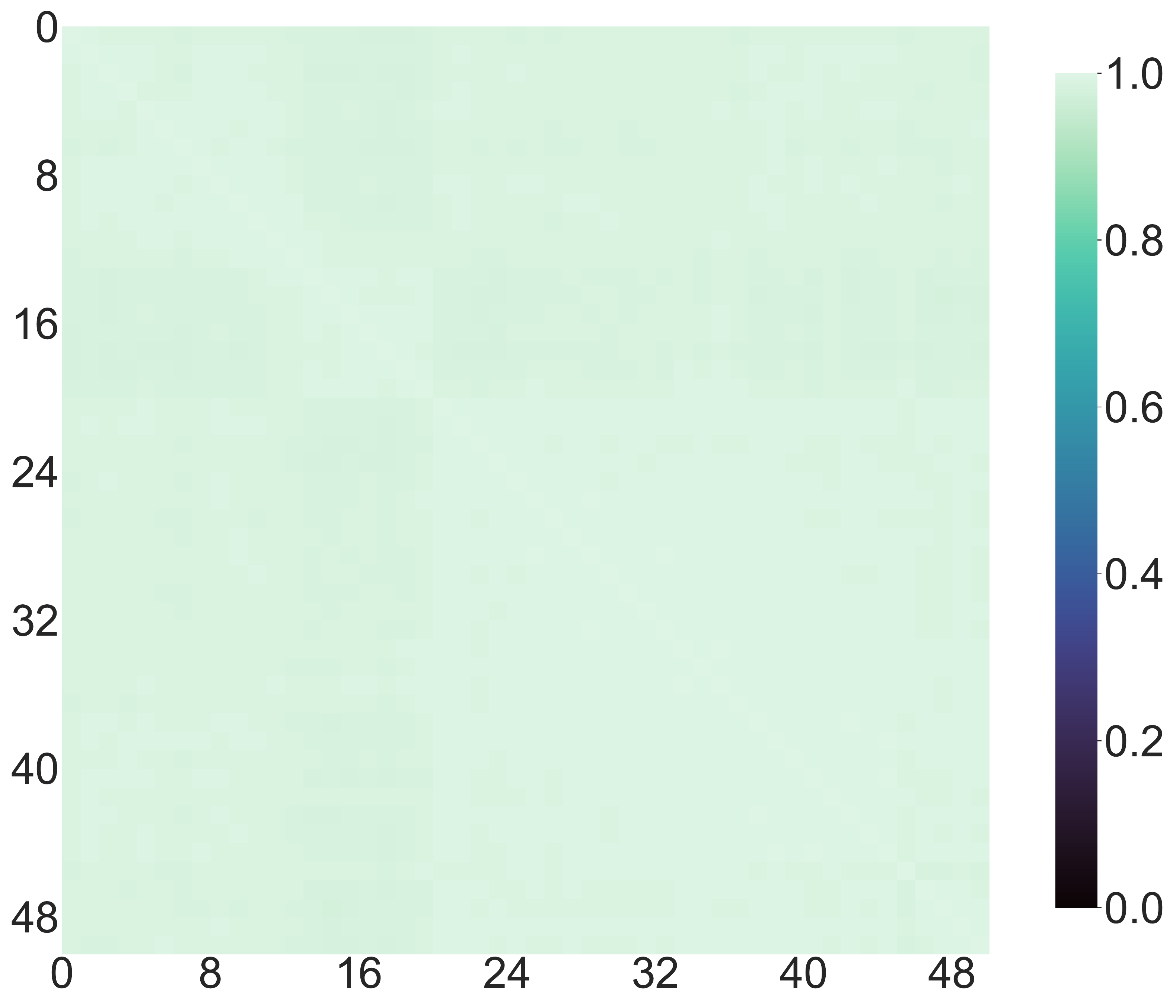}
        \caption{Task Can.}
        
    \end{subfigure}

    \caption{Similarity between features from different rollouts on Lift, Square, Tool, and Can tasks.}
    \label{fig:similarity_dlc}
\end{figure}

Figure~\ref{fig:similarity_dlc} provides additional details on rollout-level feature similarity across the Lift, Square, Tool, and Can tasks. We observe that features from different rollout iterations exhibit consistently high similarity on all tasks, reinforcing our insight that cached features from multiple directions, particularly those from rollout iterations, can be effectively leveraged to constrain large pruning errors under aggressive pruning rates.

\section{Details on Ablation Study}
For a more comprehensive evaluation, we further include ablation studies on additional tasks, namely Kitchen, Lift$_{ph}$, and the MH dataset. The results in Tables~\ref{tab: ablation_mh} and \ref{tab: ablation_kit_liftph} echo the findings from the main paper: single-direction reuse strategies consistently fail to maintain high success rates under aggressive sparsity across all tasks. In contrast, the proposed omnidirectional caching strategy consistently achieves the best trade-off between performance and efficiency across both tasks and datasets.

\begin{table}[!htbp]
  \centering
  \small
  \setlength{\tabcolsep}{3pt} 
  
  \vspace{0.5 em}
      \begin{tabularx}{\linewidth}{l *{4}{>{\centering\arraybackslash}X}}
      \toprule
      \multirow{2}{*}{\textbf{Method}} &
      \multicolumn{4}{c}{\textbf{Success Rate (\%, $\uparrow$)}} \\
      \cmidrule(lr){2-5}
      & Can & Transport & Lift & Square \\
      \midrule
      \textit{Dense}    & 94 & 56 & 100  & 74 \\
      \midrule
      Forward-direction  & 64 & 0 & 2  & 32  \\
      Timestep-direction & 72 & 48  & 98 & 66 \\
      Rollout-direction & 82 & 0 & 98 & 48 \\    
      \midrule
      \textbf{Omini-direction} & \cellcolor{best!25}{94} & \cellcolor{best!25}{62} & \cellcolor{best!25}{100} & \cellcolor{best!25}{82} \\
  \bottomrule
  \end{tabularx}
    \caption{
    \textbf{Ablation Study.}
    We report the success rate when only reusing features from single directions on MH dataset under 93\% sparsity. \colorbox{best!25}{Best} performances are highlighted.
  }
  \label{tab: ablation_mh}
\end{table}

\begin{table}[!htbp]
  \centering
  \small
  \setlength{\tabcolsep}{3pt} 

  \vspace{0.5 em}
      \begin{tabularx}{\linewidth}{l *{5}{>{\centering\arraybackslash}X}}
      \toprule
      \multirow{2}{*}{\textbf{Method}} &
      \multicolumn{5}{c}{\textbf{Success Rate (\%, $\uparrow$)}} \\
      \cmidrule(lr){2-6}
      & Kit$_{p1}$ & Kit$_{p2}$ & Kit$_{p3}$ & Kit$_{p4}$ & Lift \\
      \midrule
      \textit{Dense}    & 100 & 100 & 100  & 100 & 100\\
      \midrule
      Forward-direction  & 64 & 0 & 2  & 32  & 6\\
      Timestep-direction & 72 & 48  & 98 & 66 & 4\\
      Rollout-direction & 82 & 0 & 98 & 48 & 8\\    
      \midrule
      \textbf{Omini-direction} & \cellcolor{best!25}{100} & \cellcolor{best!25}{100} & \cellcolor{best!25}{100} & \cellcolor{best!25}{100} & \cellcolor{best!25}{100}\\
  \bottomrule
  \end{tabularx}
    \caption{
    \textbf{Ablation Study.}
    We report the success rate when only reusing features from single directions on Kitchen and Lift$_{PH}$ under 93\% sparsity. \colorbox{best!25}{Best} performances are highlighted.
  }
  \label{tab: ablation_kit_liftph}
\end{table}

\section{Training Efficiency}
\label{section: training efficiency}  
 
\begin{figure}[htbp]
    \centering 
        \centering
        \includegraphics[width=0.48\textwidth]{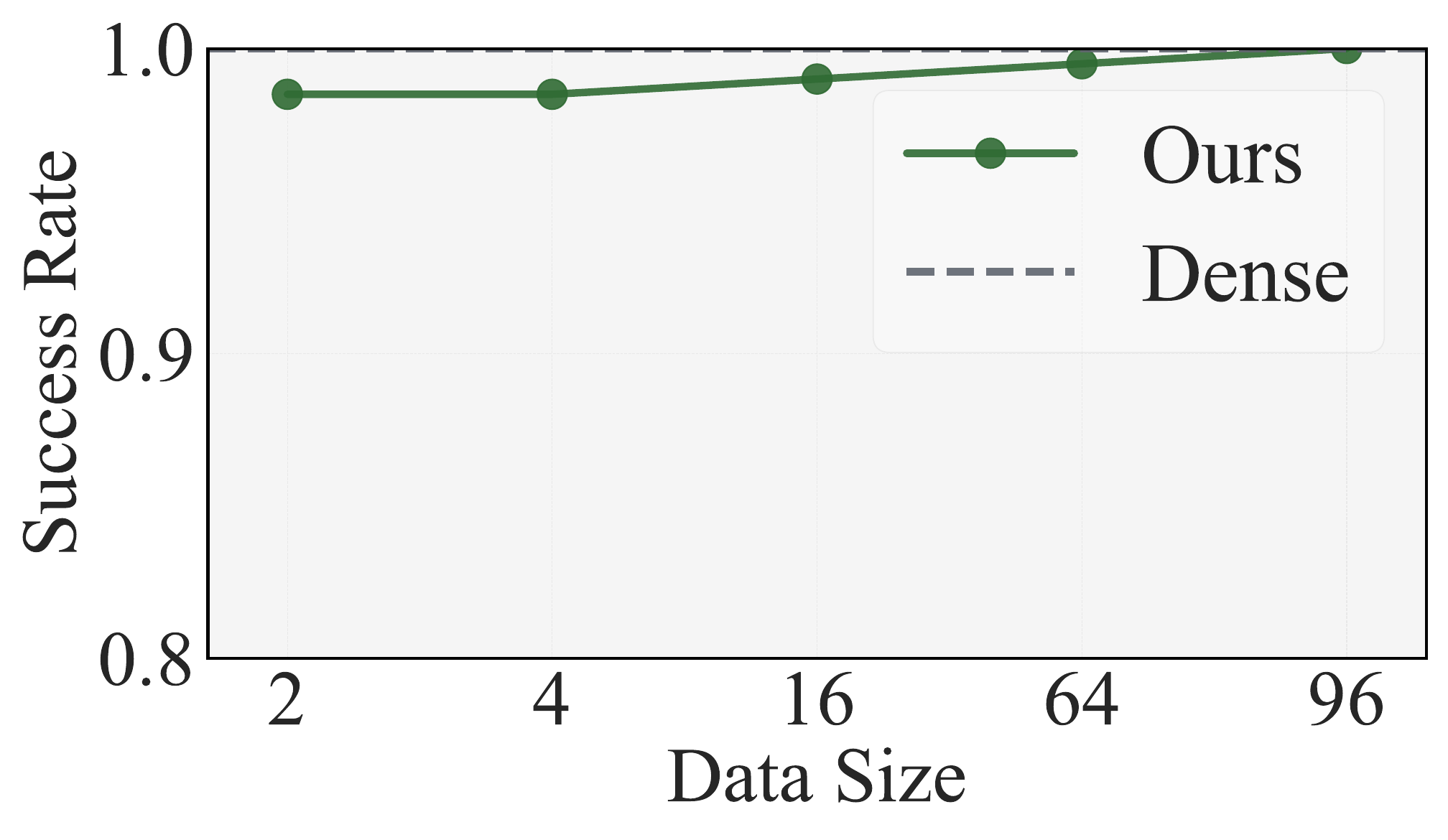}
    \caption{Success rates on the Kitchen task with different numbers of training trajectories.} 
    \label{fig:data_efficiency_kitchen} 
\end{figure}

\begin{figure}[htbp]
    \centering 
        \centering
        \includegraphics[width=0.48\textwidth]{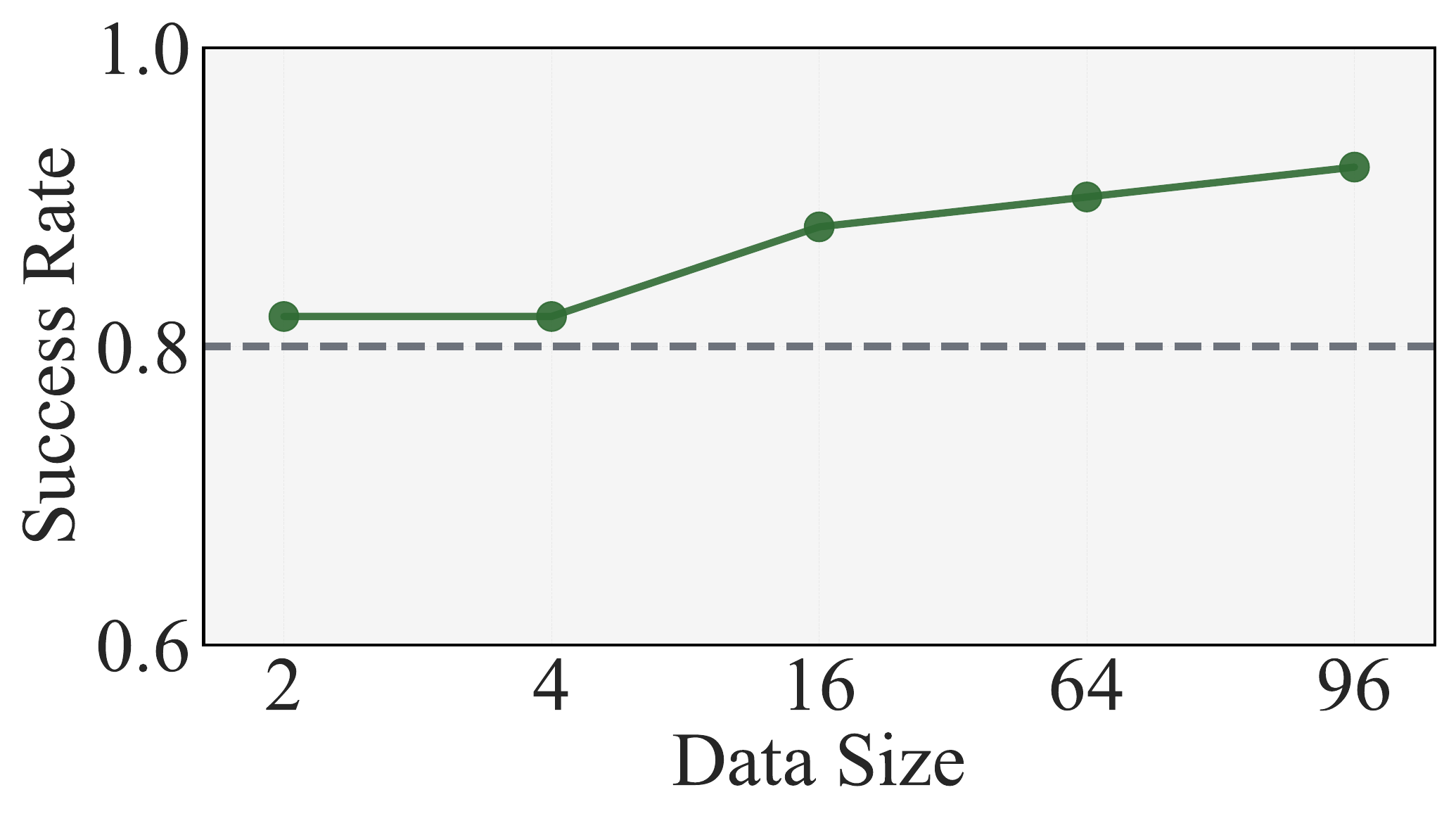}
    \caption{Success rates on the Transport task with different numbers of training trajectories.} 
    \label{fig:data_efficiency_transport} 
\end{figure}

We assess training efficiency by reporting the success rate achieved with varying numbers of sampled trajectories, as shown in Figures~\ref{fig:data_efficiency_kitchen} and~\ref{fig:data_efficiency_transport}. Remarkably, our method reaches performance comparable to the original model under 93\% sparsity using only two trajectories, and its performance continues to improve as more trajectories are provided. These results highlight the strong data efficiency of our training pipeline.

\section{Details on Qualitative Results}



\begin{figure}[htbp]
    \centering 
    
    \begin{subfigure}[b]{0.5\textwidth} 
        \centering
        \includegraphics[width=\textwidth,height=0.34\textwidth]
        {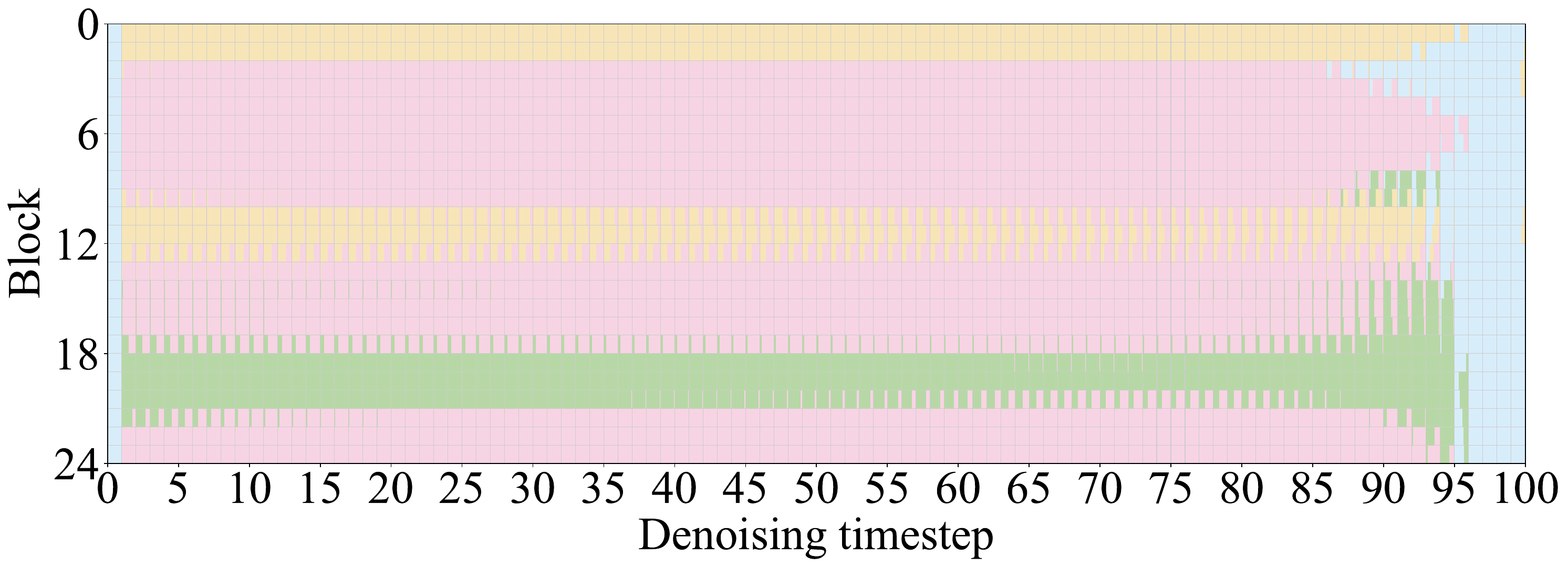}
        \caption{The 1-th Rollout Iteration}
    \end{subfigure}

    \begin{subfigure}[b]{0.5\textwidth}
        \centering
        \includegraphics[width=\textwidth,height=0.34\textwidth]{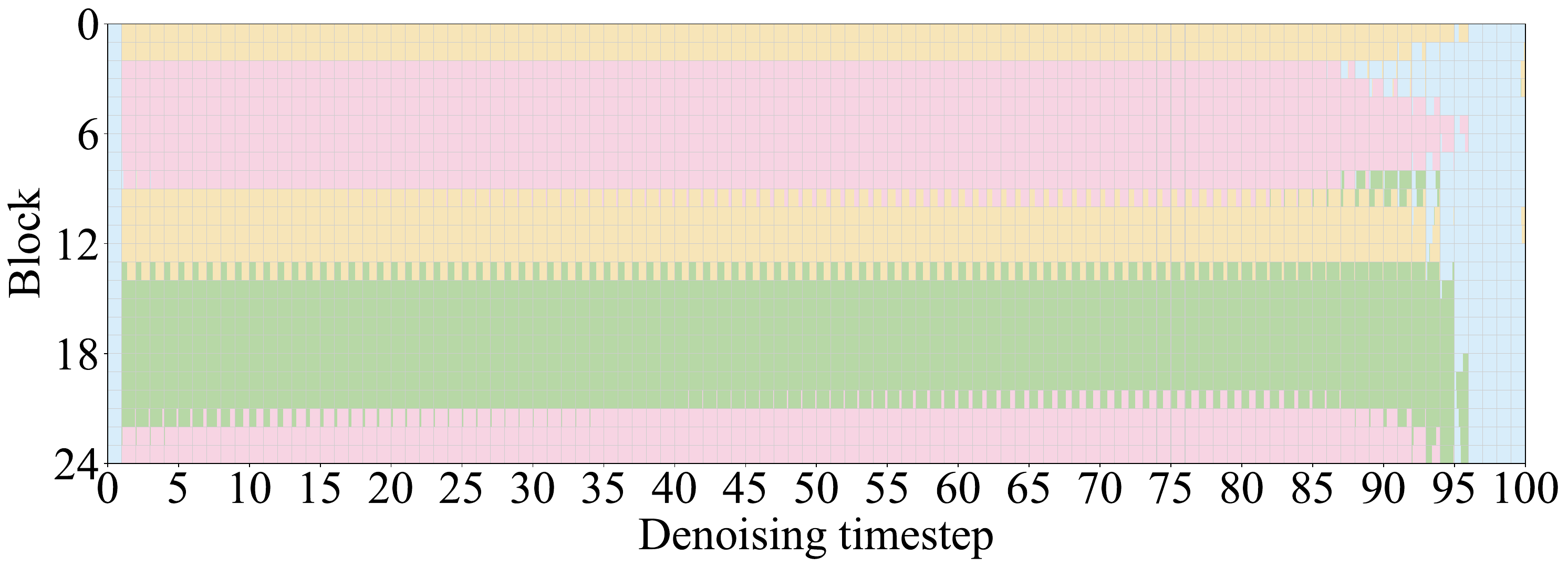}
        \caption{The 20-th Rollout Iteration}
    \end{subfigure}
    
    \begin{subfigure}[b]{0.5\textwidth}
        \centering
        \includegraphics[width=\textwidth,height=0.34\textwidth]{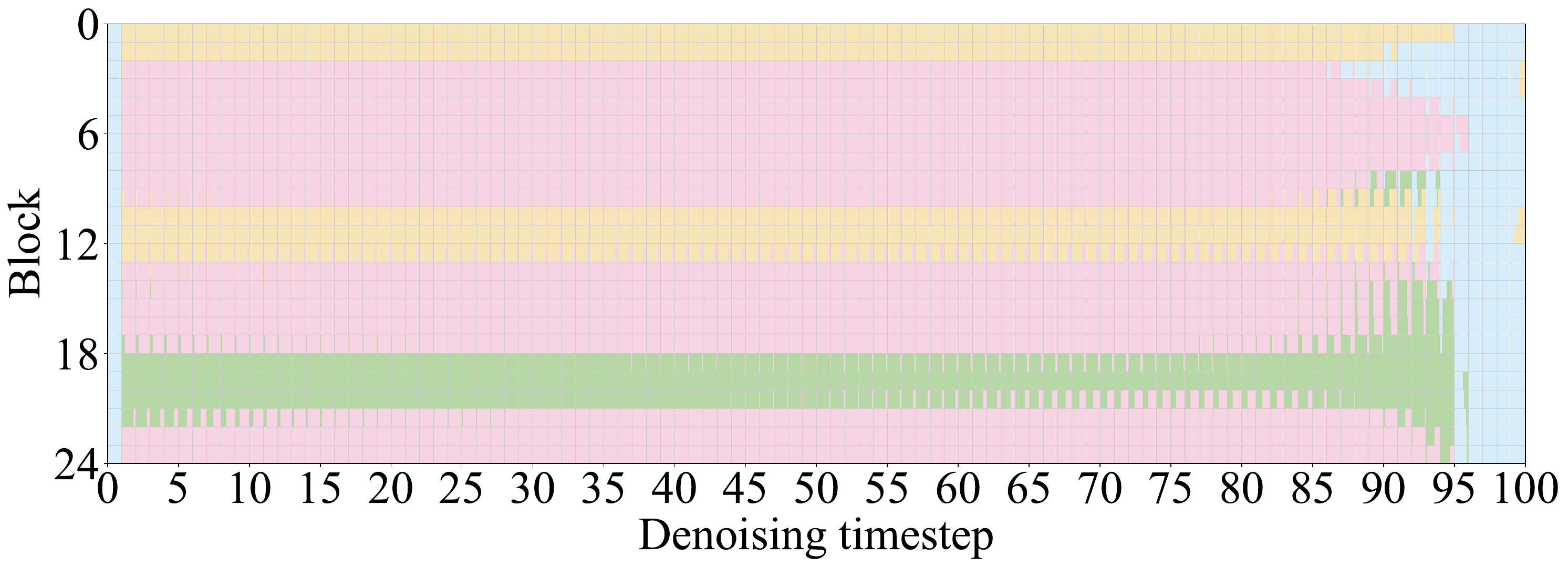}
        \caption{The 40-th Rollout Iteration}
    \end{subfigure}
    
    \caption{Visualization of $\mathcal{M}$ on Can.}
    \label{fig:cache_can} 
\end{figure}

\begin{figure}[t]
    \centering 
    
    \begin{subfigure}[b]{0.5\textwidth} 
        \centering
        \includegraphics[width=\textwidth,height=0.34\textwidth]{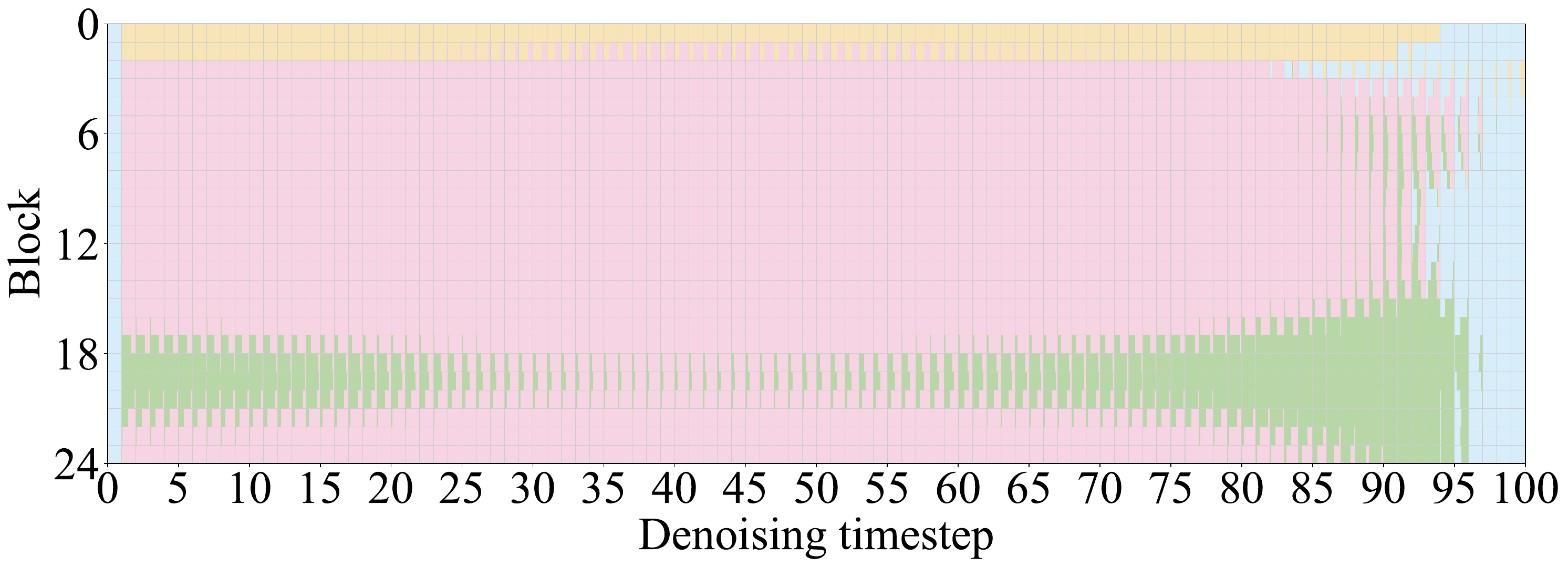}
        \caption{The 1-th Rollout Iteration}
    \end{subfigure}

    \begin{subfigure}[b]{0.5\textwidth}
        \centering
        \includegraphics[width=\textwidth,height=0.34\textwidth]{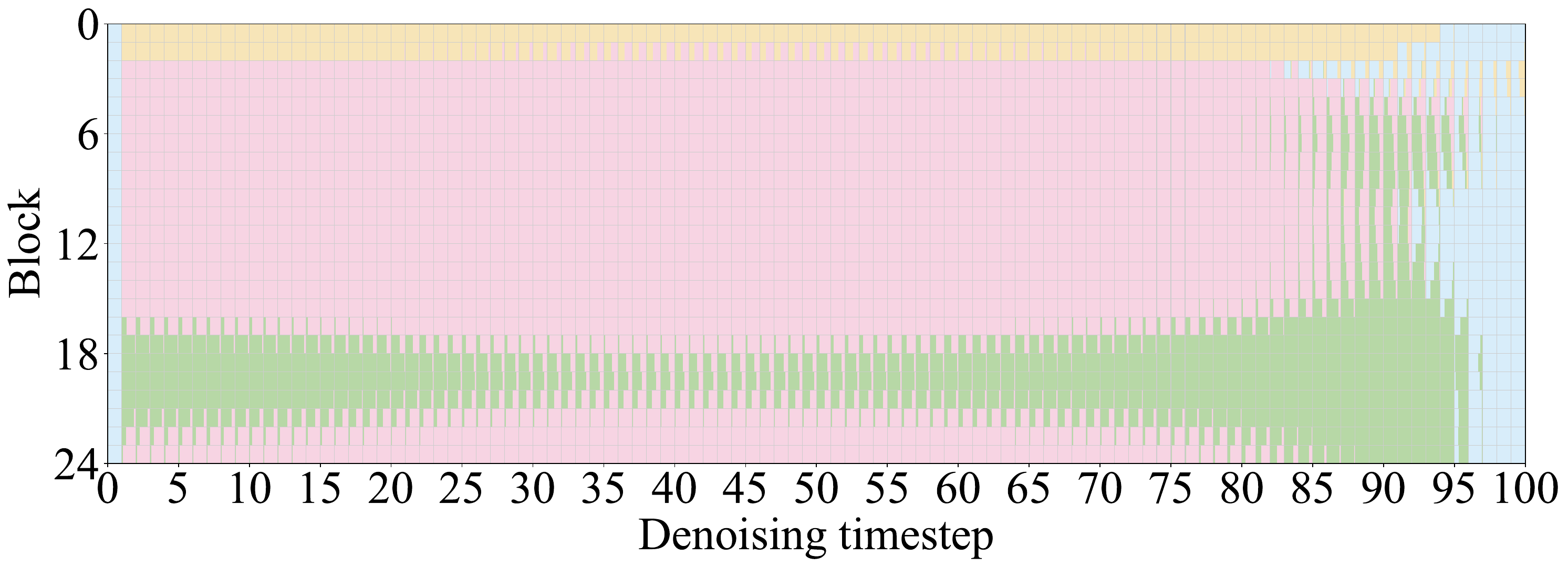}
        \caption{The 20-th Rollout Iteration}
    \end{subfigure}
    
    \begin{subfigure}[b]{0.5\textwidth}
        \centering
        \includegraphics[width=\textwidth,height=0.34\textwidth]{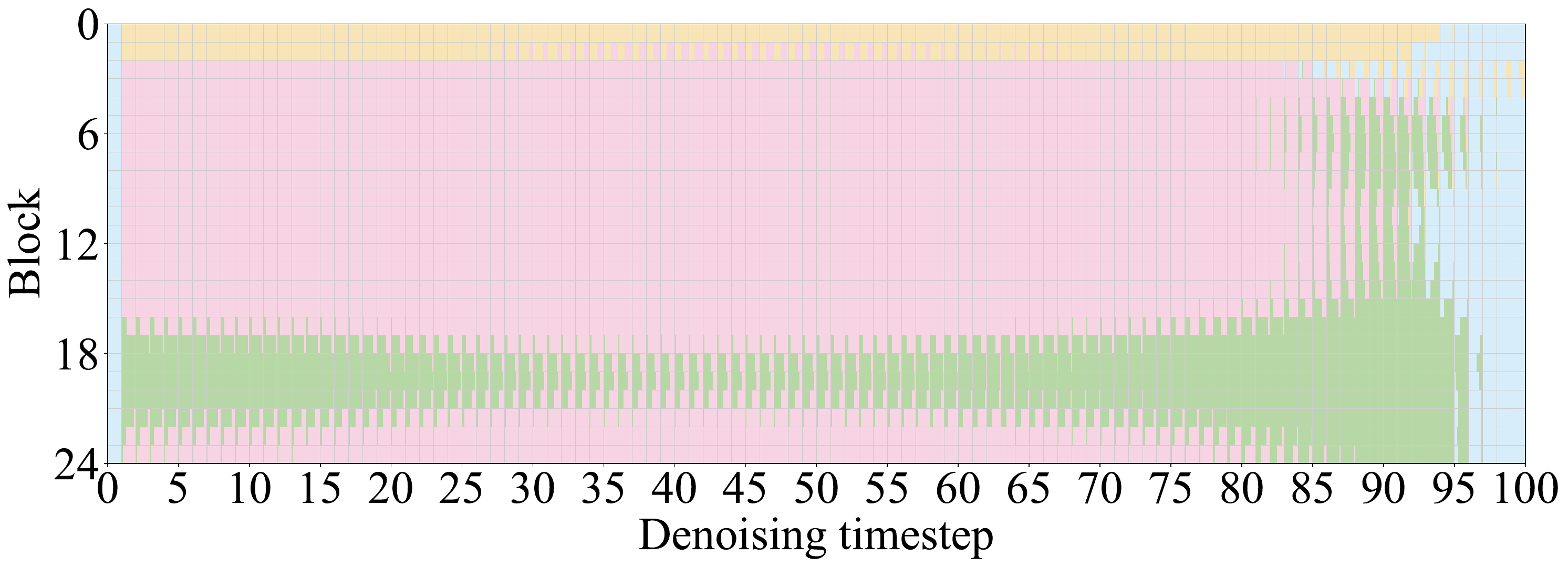}
        \caption{The 40-th Rollout Iteration}
    \end{subfigure}
    
    \caption{Visualization of $\mathcal{M}$ on Square.} 
    \label{fig:cache_square} 
\end{figure}

\begin{figure}[t]
    \centering 
    
    \begin{subfigure}[b]{0.5\textwidth} 
        \centering
        \includegraphics[width=\textwidth,height=0.34\textwidth]{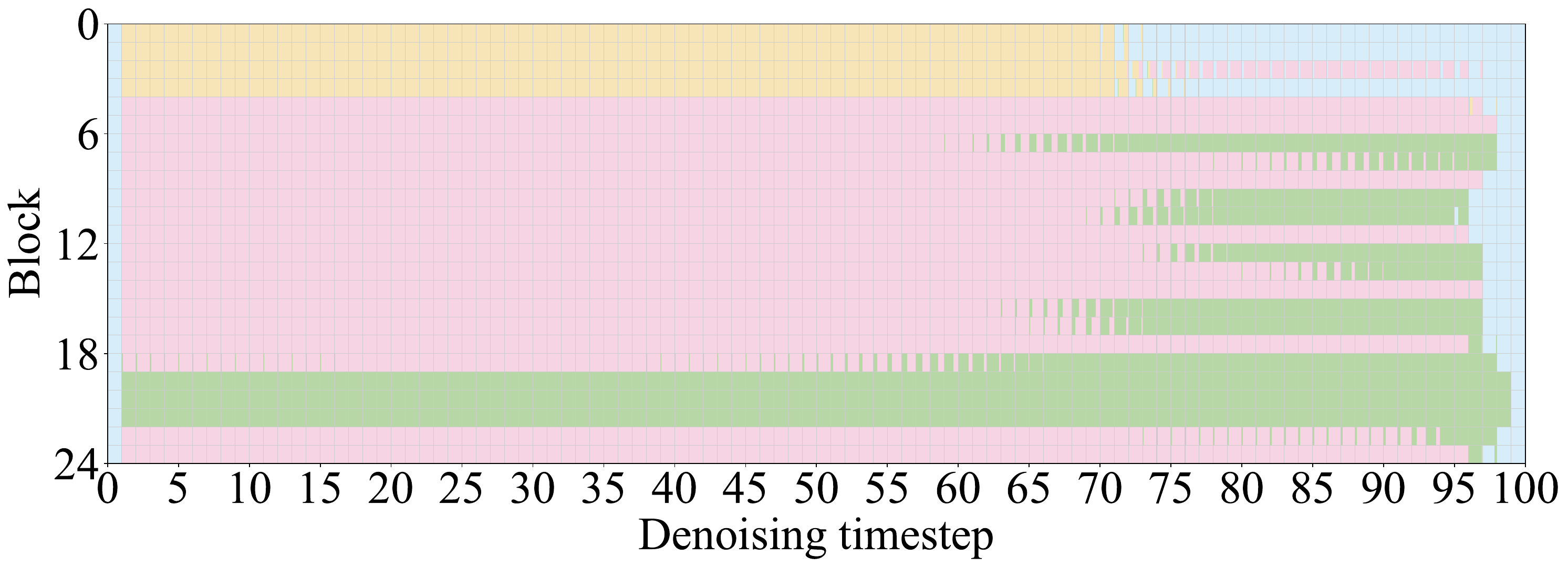}
        \caption{The 1-th Rollout Iteration}
    \end{subfigure}

    \begin{subfigure}[b]{0.5\textwidth}
        \centering
        \includegraphics[width=\textwidth,height=0.34\textwidth]{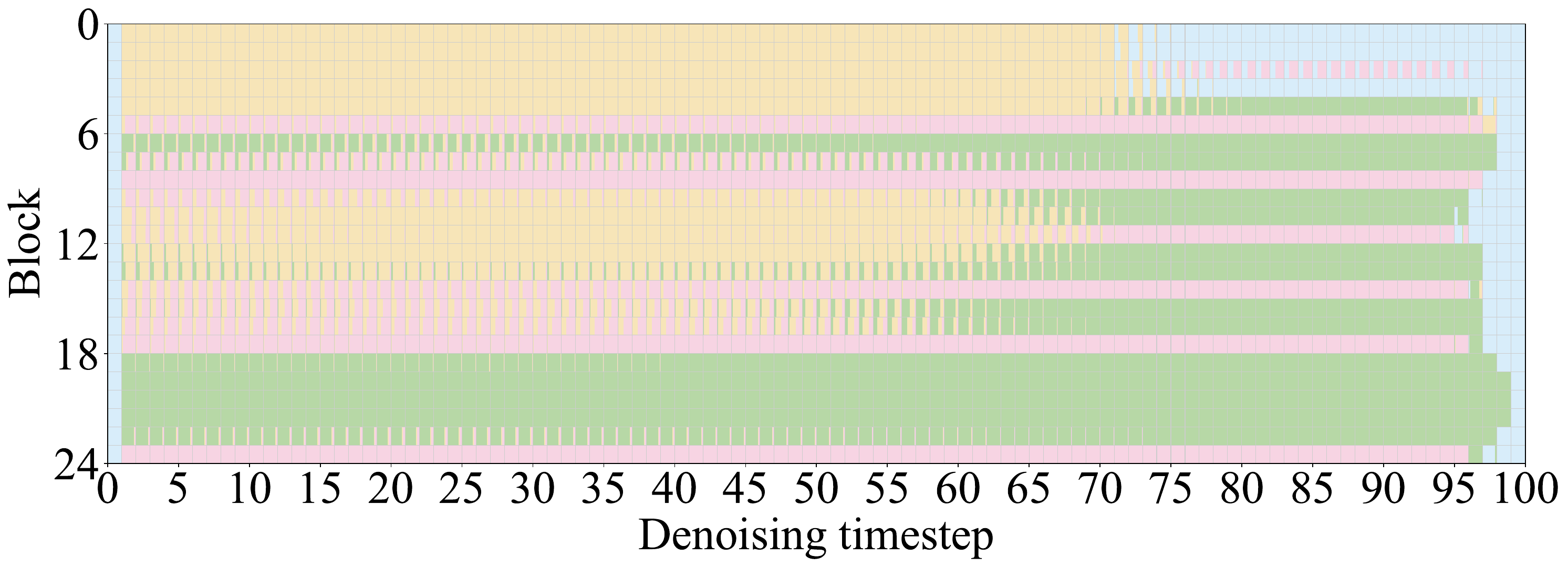}
        \caption{The 20-th Rollout Iteration}
    \end{subfigure}
    
    \begin{subfigure}[b]{0.5\textwidth}
        \centering
        \includegraphics[width=\textwidth,height=0.34\textwidth]{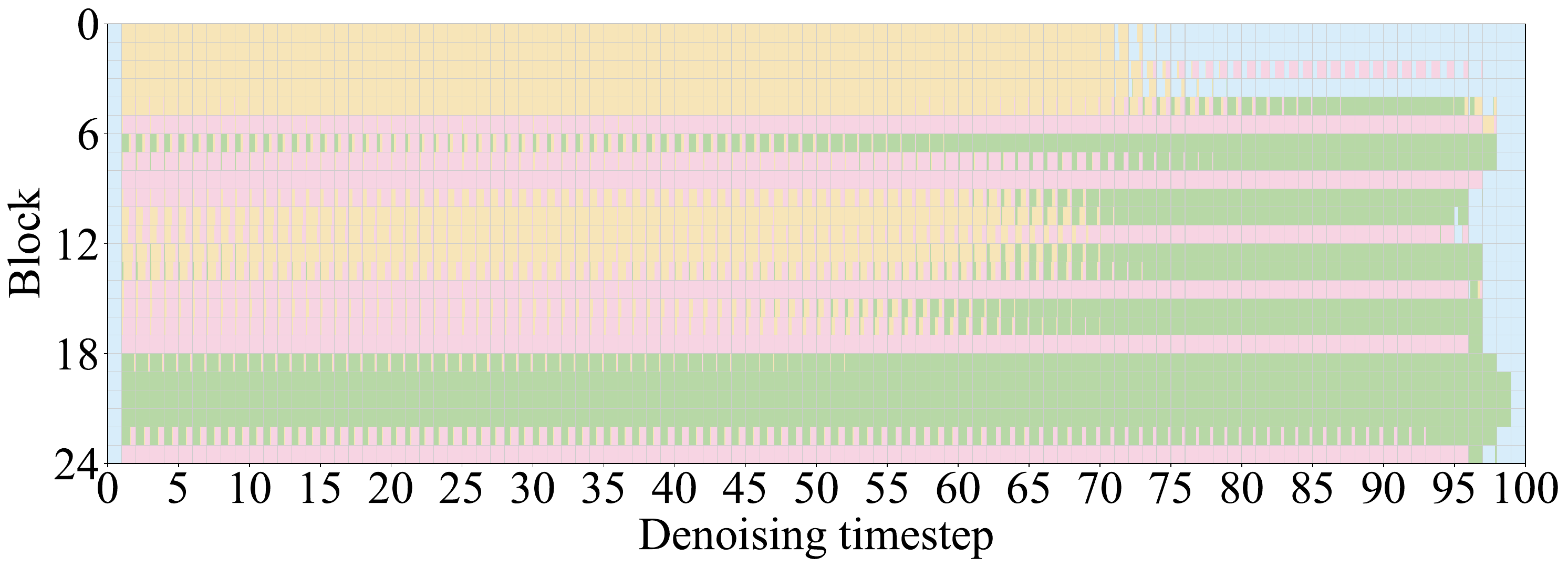}
        \caption{The 40-th Rollout Iteration}
    \end{subfigure}
    
    \caption{Visualization of $\mathcal{M}$ on Tool.}
    \label{fig:cache_tool} 
\end{figure}

\begin{figure}[t]
    \centering 
    
    \begin{subfigure}[b]{0.5\textwidth} 
        \centering
        \includegraphics[width=\textwidth,height=0.34\textwidth]{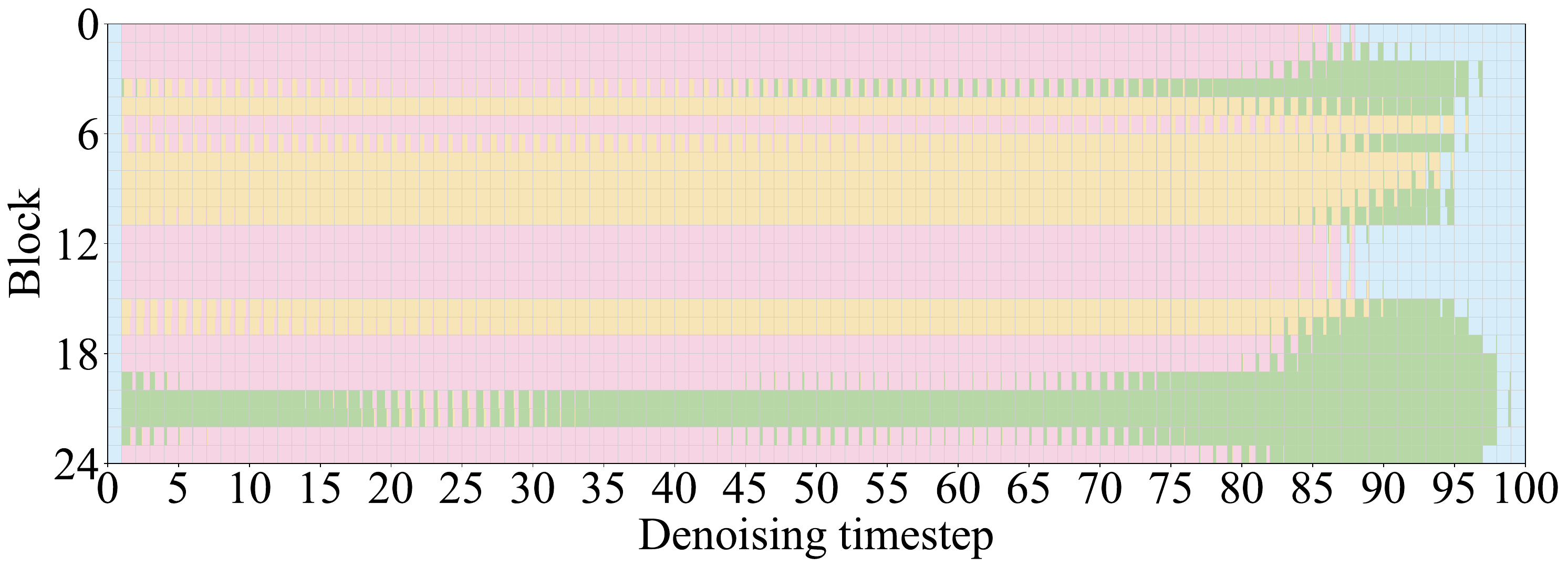}
        \caption{The 1-th Rollout Iteration}
    \end{subfigure}

    \begin{subfigure}[b]{0.5\textwidth}
        \centering
        \includegraphics[width=\textwidth,height=0.34\textwidth]{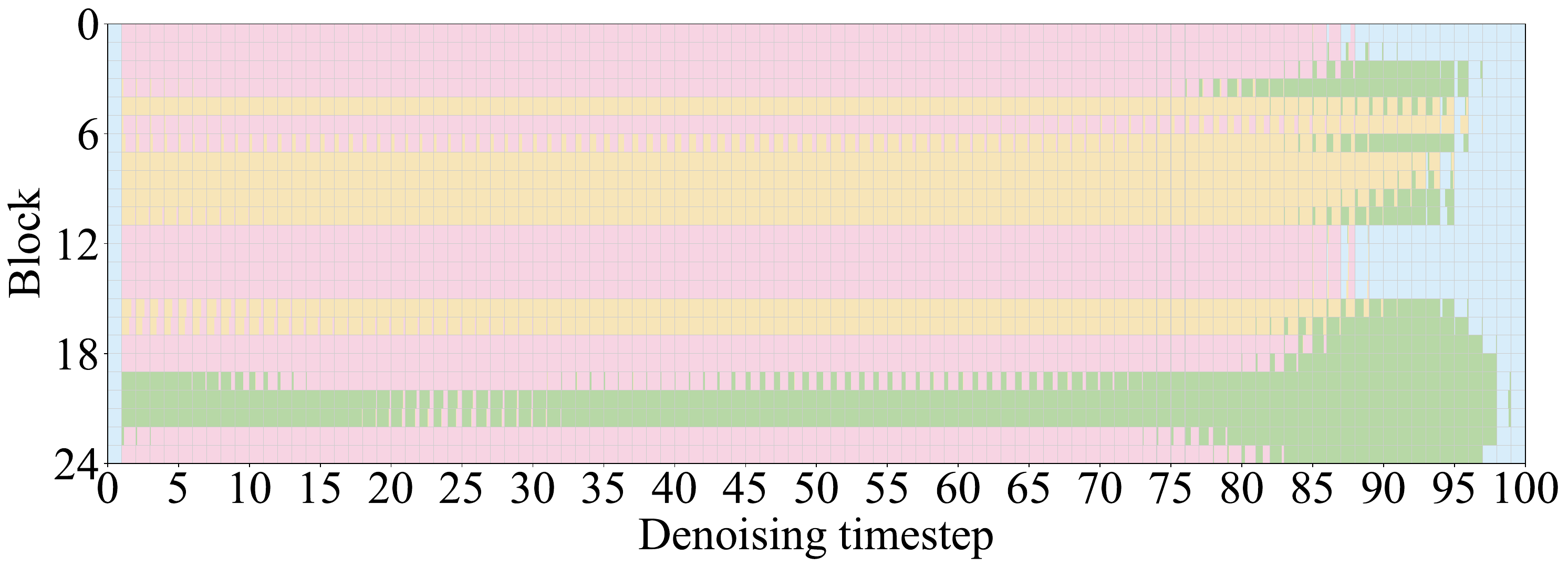}
        \caption{The 20-th Rollout Iteration}
    \end{subfigure}
    
    \begin{subfigure}[b]{0.5\textwidth}
        \centering
        \includegraphics[width=\textwidth,height=0.34\textwidth]{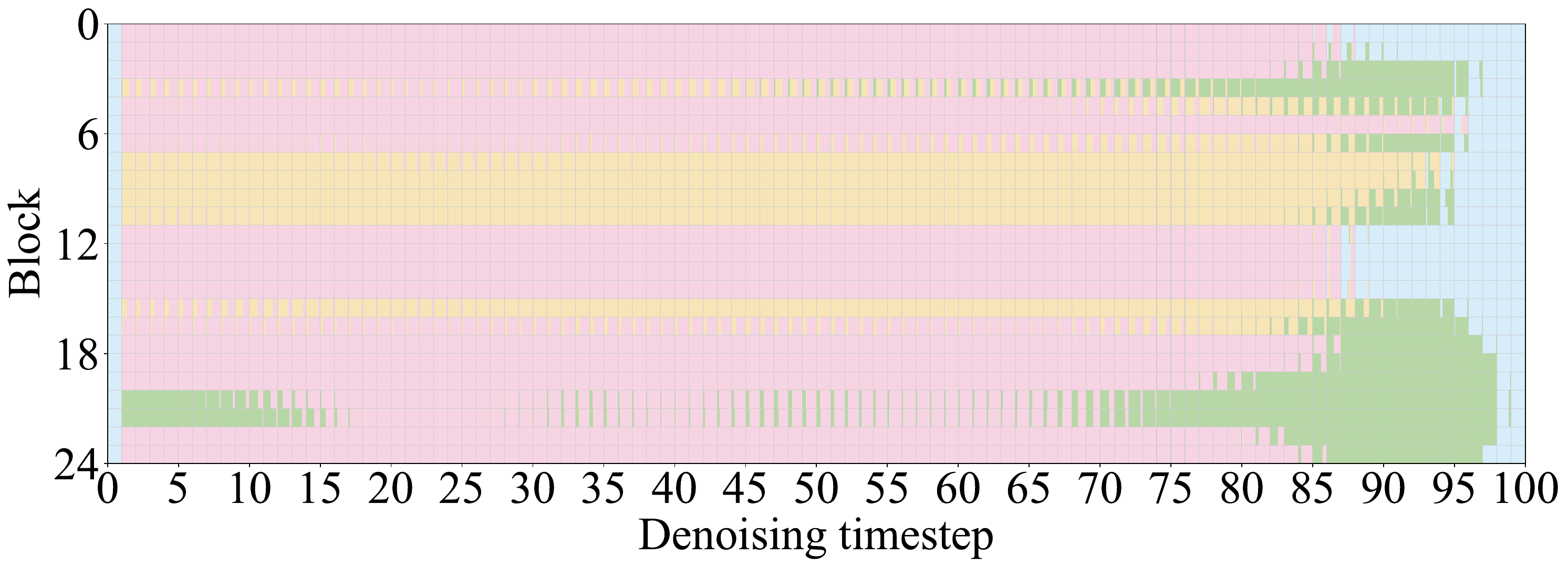}
        \caption{The 40-th Rollout Iteration}
    \end{subfigure}
    
    \caption{Visualization of $\mathcal{M}$ on Transport.} 
    \label{fig:cache_transport} 
\end{figure}

\begin{figure}[t]
    \centering 
    
    \begin{subfigure}[b]{0.5\textwidth} 
        \centering
        \includegraphics[width=\textwidth,height=0.34\textwidth]{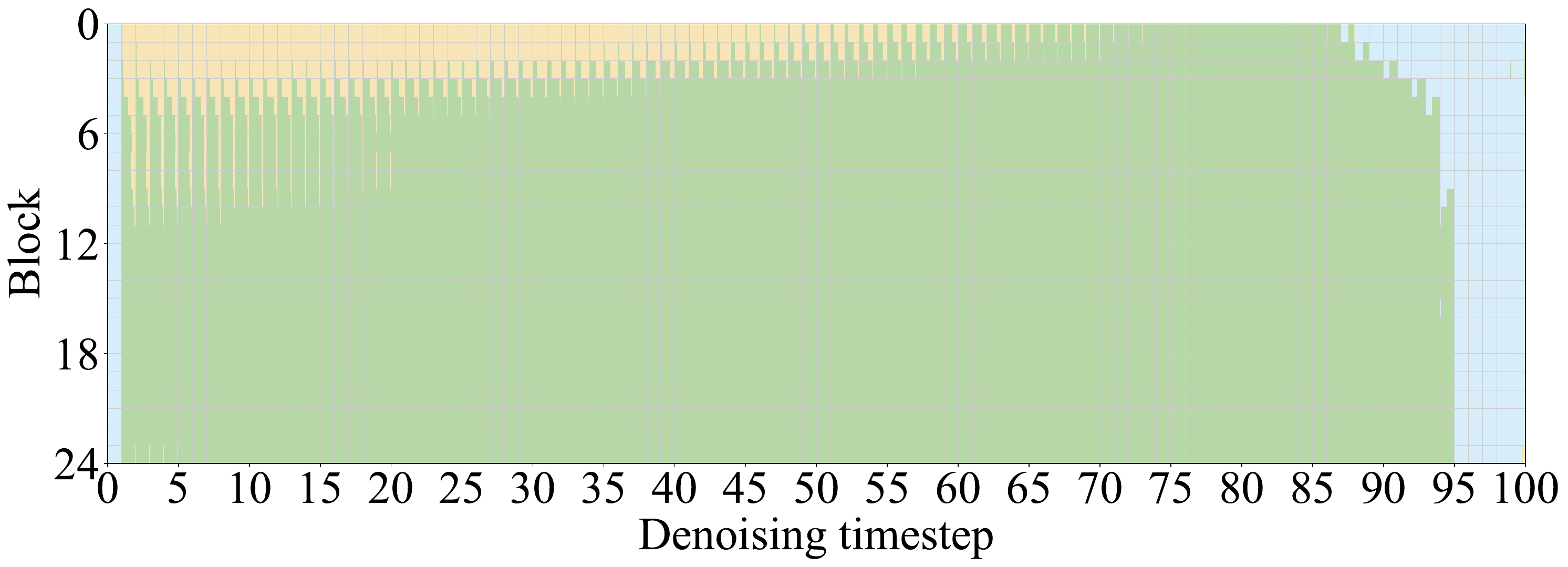}
        \caption{The 1-th Rollout Iteration}
    \end{subfigure}

    \begin{subfigure}[b]{0.5\textwidth}
        \centering
        \includegraphics[width=\textwidth,height=0.34\textwidth]{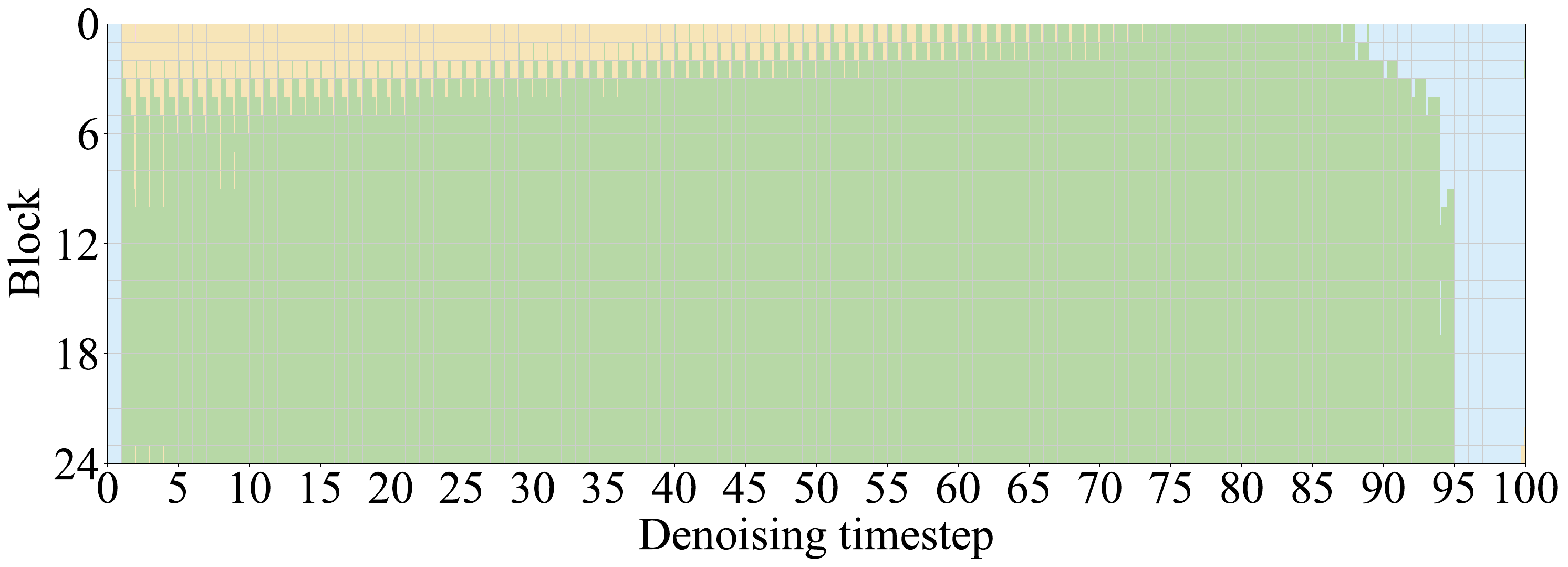}
        \caption{The 20-th Rollout Iteration}
    \end{subfigure}
    
    \begin{subfigure}[b]{0.5\textwidth}
        \centering
        \includegraphics[width=\textwidth,height=0.34\textwidth]{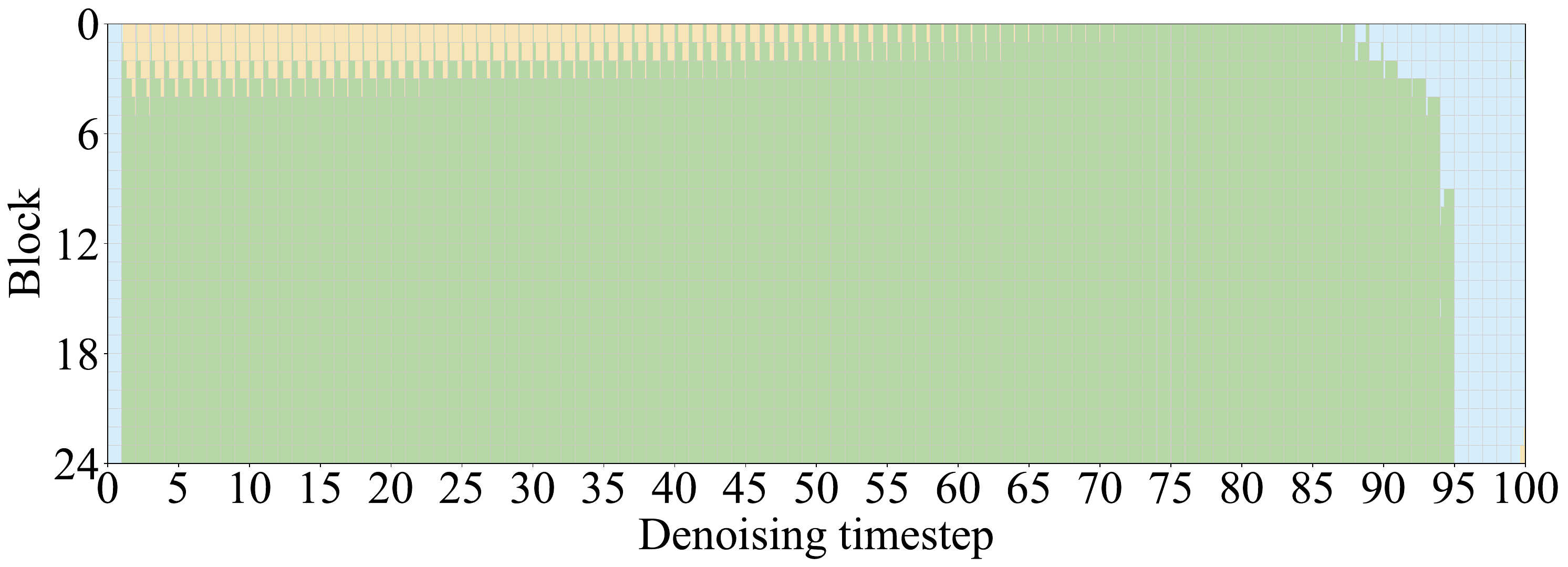}
        \caption{The 40-th Rollout Iteration}
    \end{subfigure}
    
    \caption{Visualization of $\mathcal{M}$ on Kitchen.}
    \label{fig:cache_kitchen} 
\end{figure}

We present the predicted pruning masks $\mathcal{M}$ for different tasks in Figures~\ref{fig:cache_can}, \ref{fig:cache_square}, \ref{fig:cache_tool}, \ref{fig:cache_transport}, and \ref{fig:cache_kitchen}. These visualizations highlight several key observations.

First, the pruning masks differ significantly between tasks. For example, features cached from the forward direction are prioritized in the Kitchen task, whereas features from the rollout direction dominate in the Square task. This verifies that model dynamics evolve based on distinct perceptions in an open environment.

Second, a dominant trend across all tasks is that features cached from the rollout iteration, colored in pink, account for the vast majority of preserved computations. Conversely, the previously adopted temporal-level cached features, colored in yellow, account for a minimal portion. This observation strongly suggests that the rollout direction provides a more effective caching strategy for action diffusion.

Finally, the mask $\mathcal{M}$ exhibits significant variations throughout the rollout iterations, revealing evolving visuomotor dynamics within multi-round interactions. We observe other interesting phenomena, such as computations being heavily concentrated in the final steps of the rollout. We hope these observations inspire future research into efficient visuomotor model dynamics.


\label{sec: qualitative results on can}

\label{sec: qualitative results on square}

\label{sec: qualitative results on tool}

\label{sec: qualitative results on transport}

\label{sec: qualitative results on kitchen}


\section{Generalization to Image Generation} 
Our design is tailored to the distinct characteristics of action diffusion, specifically the heavy visuomotor conditioning costs and the unexploited redundancy in long-horizon rollouts. However, the underlying principles apply broadly. The Parallelized Inference Pipeline is compatible with standard DiT-style backbones used in image/video diffusion. Moreover, Omnidirectional Reusing Strategy naturally extends to autoregressive video models (e.g., \textit{Rolling Forcing}). It can also adapt to generic image/video generation via a variant that selectively utilizes block and timestep dimensions. Table~\ref{tab:imagegen} demonstrates this strong generalization, showing negligible degradation (FID 2.23) at 80\% sparsity.

\begin{table}[!htbp]
    \centering
    \small
    \setlength{\tabcolsep}{1.5pt} 
    \renewcommand{\arraystretch}{1.18} 
    \caption{Class-to-image generation on ImageNet with DiT-XL/2.}
    \label{tab:imagegen}
    \vspace{-0.3em}
    \begin{tabular}{lcccc}
        \toprule
        \textbf{Method} & \textbf{Sparsity (\%)} & \textbf{FID($\downarrow$)} & \textbf{sFID($\downarrow$)} & \textbf{Speedup($\uparrow$)} \\
        \midrule
        DDIM 50 steps           & 0   & 2.18  & 4.21 & 1.0$\times$ \\
        \midrule
        FORA ($\mathcal{N}=5$)  & 80  & 6.58  & 11.29 & 2.87$\times$ \\
        ToCa ($\mathcal{N}=6$)  & 83  & 6.55  & 7.10  & 2.63$\times$ \\
        DuCa ($\mathcal{N}=5$)  & 80  & 6.06  & 6.72  & 2.78$\times$ \\
        TaylorSeer ($\mathcal{N}=5$) & 80 & 2.65 & 5.36 & 2.38$\times$ \\
        \midrule
        \textbf{Ours ($\rho = 80\%$)} 
            & \cellcolor{best!25} 80
            & \cellcolor{best!25} \textbf{2.23}
            & \cellcolor{best!25} \textbf{4.37}
            & \cellcolor{best!25} 2.63$\times$ \\
        \bottomrule
    \end{tabular}
\end{table}

\end{document}